\documentclass[10pt,twocolumn,letterpaper]{article}

\usepackage{cvpr}
\usepackage{times}
\usepackage{epsfig}
\usepackage{graphicx}
\usepackage{subfigure}
\usepackage{amsmath}
\usepackage{amssymb}

\usepackage[pagebackref=true,breaklinks=true,letterpaper=true,colorlinks,bookmarks=false]{hyperref}

\cvprfinalcopy 

\def\cvprPaperID{1710} 

\ifcvprfinal\fi
\begin{document}

\title{Deep Semantic Ranking Based Hashing for Multi-Label Image Retrieval}

\author{Fang Zhao \quad
Yongzhen Huang \quad
Liang Wang \quad
Tieniu Tan \\
Center for Research on Intelligent Perception and Computing\\
Institute of Automation, Chinese Academy of Sciences\\
{\tt\small \{fang.zhao,yzhuang,wangliang,tnt\}@nlpr.ia.ac.cn}
}

\maketitle
\thispagestyle{empty}

\begin{abstract}
With the rapid growth of web images, hashing has received increasing interests in large scale image retrieval. Research efforts have been devoted to learning compact binary codes that preserve semantic similarity based on labels. However, most of these hashing methods are designed to handle simple binary similarity. The complex multilevel semantic structure of images associated with multiple labels have not yet been well explored. Here we propose a deep semantic ranking based method for learning hash functions that preserve multilevel semantic similarity between multi-label images. In our approach, deep convolutional neural network is incorporated into hash functions to jointly learn feature representations and mappings from them to hash codes, which avoids the limitation of semantic representation power of hand-crafted features. Meanwhile, a ranking list that encodes the multilevel similarity information is employed to guide the learning of such deep hash functions. An effective scheme based on surrogate loss is used to solve the intractable optimization problem of nonsmooth and multivariate ranking measures involved in the learning procedure. Experimental results show the superiority of our proposed approach over several state-of-the-art hashing methods in term of ranking evaluation metrics when tested on multi-label image datasets.
\end{abstract}

\section{Introduction}

Representing images efficiently is an important task for large scale content-based image retrieval. Binary hashing has attracted extensive attention due to computational and storage efficiencies of binary hash codes. It aims to map high-dimensional image data to compact binary codes in a Hamming space while maintaining some notion of similarity (e.g., metric similarity in the original feature space or semantic similarity based on labels).

\begin{figure}[t]
  \centering
  \includegraphics[width=8.0cm]{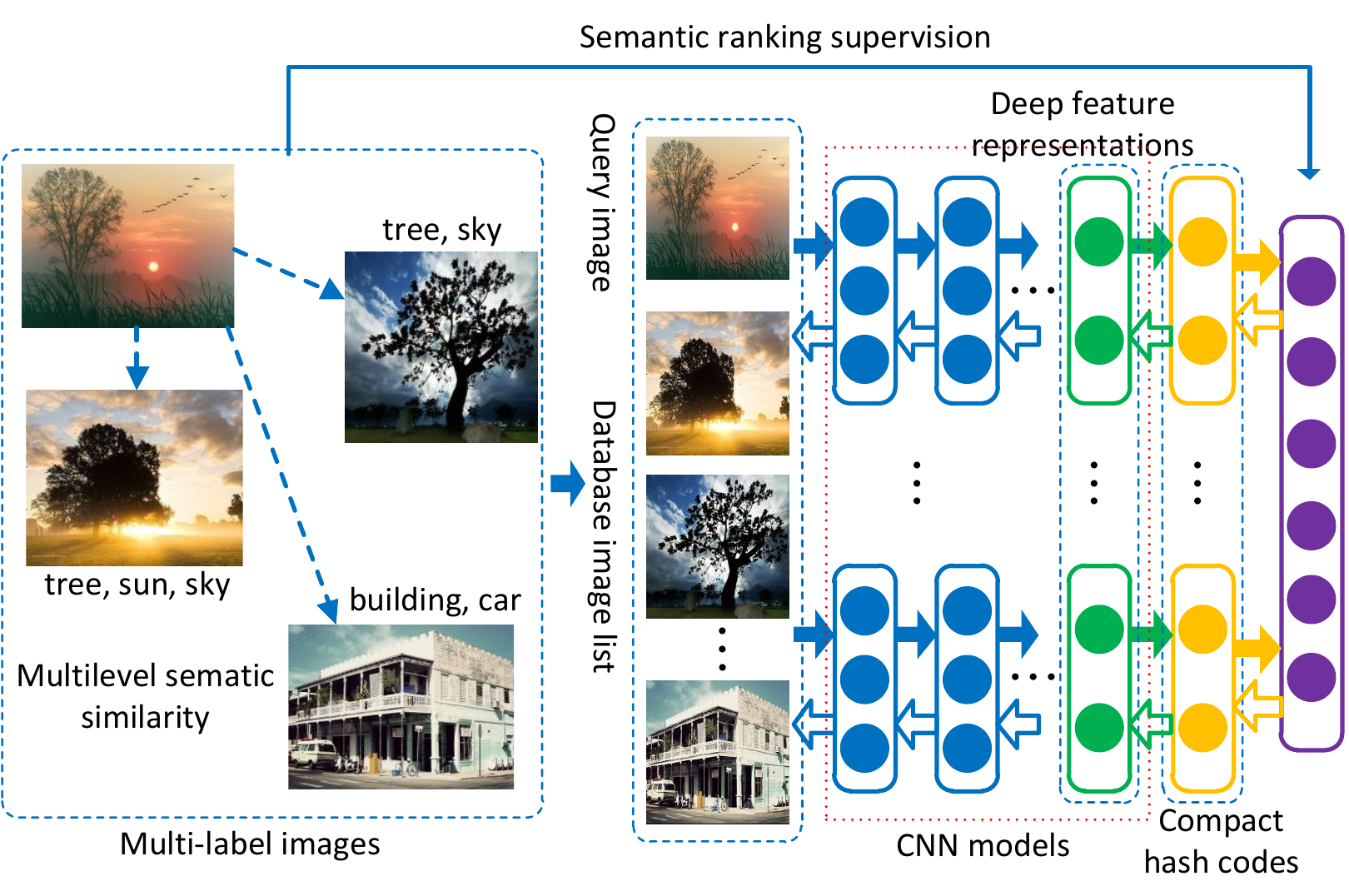}
  \caption{The proposed deep semantic ranking based hashing. Solid and hollow arrows indicate forward and backward propagation directions of features and gradients respectively. Hash functions consist of deep convolutional neural network (CNN) and binary mappings of the feature representation from the top hidden layers of CNN. Multilevel semantic ranking information is used to learn such deep hash functions to preserve the semantic structure of multi-label images.} \label{fig_1}
\end{figure}

Early hashing methods are data-independent, such as locality sensitive hashing \cite{R1} and its variants, which use random projections as hash functions without exploring the data distribution. Recently, various data-dependent hashing methods have been proposed, which learn hash functions according to the data distribution. Some of them mainly focus on preserving the metric structure of the data in the original feature space, such as spectral hashing \cite{R2} and binary reconstructive embedding \cite{R3}. However, distance metrics (e.g., Euclidean distance) in the original space sometimes cannot measure well the semantic similarity that is essential for image retrieval.

To preserve semantic structure of the data, hashing methods with supervisory information in form of class labels have been further developed \cite{R8,R4,R6,R7,R33}. Through formulating hash function learning as a classification problem or as an optimization problem of pairwise relation based loss functions, these methods are able to learn hash codes which preserve binary semantic similarity. But in practice images are usually simultaneously associated with multiple semantic labels, and in this case the similarity relationship between two images is more complex and is usually relevant to the number of common labels that images have. Consequently, a multilevel measure (such as very similar, normally similar and dissimilar) is required to describe the similarity, which cannot be handled well by the above methods and has not been studied well.

Besides, the learning capability of the standard pipeline followed by most hashing method, i.e., firstly extracting features like GIST \cite{R28} and SIFT \cite{R29} as image representations, and then learning mappings from these representations to binary codes, is inadequate for dealing with relatively complex semantic structure due to the semantic information loss in the hand-crafted features. Thus more effective semantic feature representation is also desirable.

In this paper, we introduce a novel framework based on semantic ranking and deep learning model for learning hash functions that preserve multilevel similarity between multi-label images in the semantic space. An overall view of the proposed framework termed deep semantic ranking based hashing (DSRH) is illustrated in Fig. \ref{fig_1}. Here we use deep convolutional neural network (CNN) \cite{R15} to construct hash functions to learn directly from images, which provides much richer sematic information than hand-crafted features. Meanwhile, we learn such deep hash functions with semantic ranking supervision which is the order of a ranking list derived from shared class labels between query and database images. The learning is a joint optimization of feature representation and mappings from them to hash codes, and it is more effective than the conventional two-stage pipeline. A ranking loss defined on a set of triplets is used as surrogate loss to solve the optimization problem resulting from nonsmooth and multivariate ranking measures, and then the stochastic gradient descent algorithm can be used to optimize model parameters. We evaluate the proposed DSRH method on a couple of multi-label image datasets and compare it with several state-of-the-art hashing methods based on both hand-crafted features and activation features from the CNN model. Experimental results demonstrate that our method is able to capture complex multilevel semantic structure and significantly outperforms other hashing methods in ranking quality.

Our main contributions include: 1) A novel hash function learning formwork is proposed to combine semantic ranking and deep learning model to address the problem of preserving multilevel semantic similarity between multi-label images. To the best of our knowledge, it is the first time to exploit deep convolutional neural network with listwise ranking supervision for hashing. 2) A scheme based on surrogate losses is applied to the proposed framework to effectively solve the optimization problem of ranking measures. 3) Our method boosts the benchmark of multi-label image retrieval, achieving the state-of-the-art performance in terms of ranking evaluation metrics.

The rest of this paper is organized as follows. Related work is briefly discussed in Section 2. The proposed deep semantic ranking based hashing is formulated and optimized in Section 3. Experimental evaluations are presented in Section 4. Finally, Section 5 concludes this paper.

\section{Related Work}

As described before, the existing hash methods can be roughly divided into two categories: data-independent and data-dependent. Here we mainly discuss data-dependent hash methods preserving the semantic structure which this paper focuses on. Iterative quantization with canonical correlation analysis (CCA-ITQ) \cite{R5} utilizes CCA with labels to reduce the dimensionality of input data and binarizes the outcome through minimizing the quantization error, where only the pointwise label information is exploited to guide hash function learning. By comparison, some approaches try to preserve the semantic similarity based on pairwise relation. Boosted similarity sensitive coding (BSSC) \cite{R8} assigns each pair of data points a label to learn a set of weak classifiers as hash functions. Semi-supervised hashing (SSH) \cite{R6} minimizes an empirical error over the labeled pairs of points and makes hash codes balanced and uncorrelated to avoid overfitting. Motivated by latent structural SVM, minimal loss hashing (MLH) \cite{R7} proposes a pairwise hinge-like loss function and minimizes its upper bound to learn similarity-preserving binary codes.

Furthermore, order-preserving approaches, which are more related to this paper, explicitly use ranking information in objective functions to learn hash codes that preserve the similarity order in the feature or semantic space. Order preserving hashing (OPH) \cite{R9} formulates an alignment between the similarity orders computed respectively from the original Euclidean space and the Hamming space, which can be solved using the quadratic penalty algorithm. On the basis of \cite{R7}, hamming distance metric learning (HDML) \cite{R10} develops a metric learning framework based on a triplet ranking loss to preserve relative similarity. However, this triplet loss function only considers local ranking information and is limited in capturing information about multilevel similarity. By using a triplet representation for listwise supervision, ranking-based supervised hashing (RSH) \cite{R11} minimizes the inconsistency of ranking order between the hamming and original spaces to keep global ranking order. Different from RSH, our method leverages deep learning model to discover deeper semantic similarity and can scale well on large training sets. Column generation hashing (CGH) \cite{R12} and StructHash \cite{R14} combine ranking information with the boosting framework to learn a weighted hamming embedding. In contrast, our method needs no extra weight to rank hash codes.

Deep learning models, particularly deep convolutional neural networks (CNNs), have achieved great success in various visual tasks such as image classification, annotation, retrieval and object detection \cite{R15,R31,R32,R16,R17} due to their powerful representation learning capability. Some ranking loss based CNNs have been explored in these tasks. Wang et al. \cite{R16} use a ranking loss based on triplet sampling in CNNs to learn image similarity metric. Gong et al. \cite{R31} incorporate a warp approximate ranking into CNNs for image annotation. There are a few hashing methods that also use deep models. Salakhutdinov et al. \cite{R34} use a deep generative model as hash functions. Similarly, Torralba et al. \cite{R4} model a deep network by using multiple layers of RBMs. Given approximate hash codes learned from pairwise similarity matrix decomposition, Xia et al. \cite{R19} learn hash functions using CNNs to fit the learned hash codes. However, these methods do not explicitly impose the ranking constraint on the deep models, which can not figure out the multi-level similarity problem.


\section{Our Method}

In general, a hash function $h:{\mathbb{R}^D} \to \{ -1,1\}$ is treated as a mapping that projects a \emph{D}-dimensional input onto a binary code. Assume that we are given a set of class labels $\mathcal{L} = \{ 1,...,C\}$ and a dataset $\mathcal{D} = \{ {\mathbf{x}}_n\} _{n = 1}^N$ where each data point ${\mathbf{x}} \in {\mathbb{R}^D}$ is associated with a subset of labels $\mathcal{Y} \subseteq \mathcal{L}$, our goal is to learn a set of hash functions ${\mathbf{h}}({\mathbf{x}}) = \left[ {{h_1}({\mathbf{x}}),{h_2}({\mathbf{x}}),...,{h_K}({\mathbf{x}})} \right]$ that generates \emph{K}-bit ($K \ll D$) binary codes while preserving the semantic structure of data points with multiple labels.

\subsection{Deep Hash Functions}

A good form of hash functions is important for obtaining desirable hash codes. As mentioned earlier, most conventional hashing methods first extract visual features like GIST and SIFT from images and then learn ``shallow'' (usually linear) hash functions upon these features. However, these hand-crafted features have limited representation power and may lose key semantic information which is important to the task of similarity search. Here we consider designing deep hash functions using CNNs to jointly learn feature representations from raw pixels of images and their mappings to hash codes. This non-linear hierarchical hash function has more powerful learning capability than the shallow one based on features extracted in advance, and thus is able to learn feature representations more suitable for multilevel semantic similarity search.

\begin{figure}[t]
  \centering
  \includegraphics[width=8.3cm]{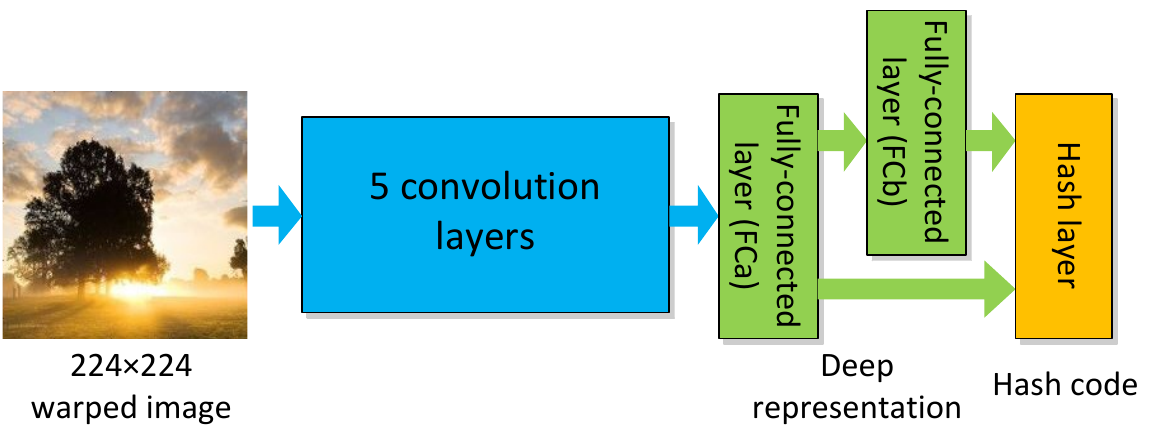}
  \caption{The structure of deep hash functions. An input image is first transformed to a fixed size, and then goes through five convolution layers and two fully-connected layers, which provides a deep feature representation. Finally, the hash layer generates a compact binary code. The hash layer is also directly connected to the first fully-connected layer (FCa) in order to utilize diverse feature information biased toward visual appearance.} \label{fig_2}
\end{figure}

As shown in Fig. \ref{fig_2}, we construct hash functions through incorporating the CNN model whose architecture is the same as \cite{R22}. A deep feature representation is computed by forward propagating a mean-subtracted $224 \times 224$ image through five convolutional layers and two fully connected layers, and then fed into the last hash layer to generate a compact binary code. Please refer to \cite{R15,R22} for more details about the geometry of the convolutional layers, local normalization and max pooling. Unlike image classification, hash codes are expected to contain global feature information within an image when used for retrieval. Thus instead of cropping an image, we warp all pixels in the image to the required size. Inspired by \cite{R23}, we add a bypassing connection between the first fully connected layer (FCa) (called the skipping layer) and the hash layer to reduce the possible information loss. We argue that the features from the second fully connected layer (FCb) of CNN are dependent on classes too much and have strong invariance, which is unfavorable for capturing subtle sematic distinction. Thus we connect the hash layer to both the two fully-connected layers to enable it encoding more diverse information biased toward visual appearance. Accordingly we define a deep hash function as:
\begin{equation}\label{5}
  h({\mathbf{x}};{\mathbf{w}}) = \operatorname{sign} ({{\mathbf{w}}^{\rm T}}[{f_a}({\mathbf{x}});{f_b}({\mathbf{x}})]),
\end{equation}
where $\mathbf{w}$ denotes weights in the hash layer, $f_a(.)$ and $f_b(.)$ denote feature vectors from the outputs of the layers FCa and FCb respectively and can be represented as the composition of the functions of previous layers. Here bias terms and parameters of $f_a(.)$ and $f_b(.)$ are omitted for the sake of concision. To obtain a \emph{K}-bit binary code, ${\mathbf{h}}({\mathbf{x}};{\mathbf{W}}) = \left[ {{h_1}({\mathbf{x}};{{\mathbf{w}}_1}),{h_2}({\mathbf{x}};{{\mathbf{w}}_2}),...,{h_K}({\mathbf{x}};{{\mathbf{w}}_K})} \right]$ can be computed.

\subsection{Semantic Ranking Supervision}

When each data point in $\mathcal{D}$ is associated with a single class label, pairs of points could be labelled as either similar or dissimilar according to whether they have the same label, and the learning procedure of hash functions is simply to make the Hamming distances between binary codes small (large) for similar (dissimilar) pairs. However, in the case of multiple labels, there exits multilevel similarity between data points depending on how many common labels they have. To preserve such multilevel semantic structure, one of the most essential ways is that for individual data points, we keep the ranking order of neighbors computed by the Hamming distance consistent with those derived from semantic labels in terms of ranking evaluation measures.

Assume we have a sample point from $\mathcal{D}$ as a query $\mathbf{q}$. For the query $\mathbf{q}$, a semantic similarity level $r$ of a database point $\mathbf{x}$ with $\mathbf{q}$ can be calculated based on the number of their common labels. The most similar database points are those sharing all the labels with $\mathbf{q}$, and assigned a level $r = \left| \mathcal{Y}_q \right|$. Accordingly, the second similar points, which share any $\left| \mathcal{Y}_q \right| - 1$ of the labels, are assigned a level $r = \left| \mathcal{Y}_q \right| - 1$. At last, the dissimilar points share none of the labels and are assigned a level $r = 0$. And then we can obtain a ground-truth ranking list for $\mathbf{q}$ by sorting the database points in decreasing order of their similarity levels. According to the ground-truth ranking, various evaluation criteria can be used to measure the consistency of the rankings predicted by hash functions, such as the Normalized Discounted Cumulative Gain (NDCG) score \cite{R21}, which is a popular measure in the information retrieval community and defined as:
\begin{equation}\label{1}
  NDCG@p = \frac{1}{Z}\sum\limits_{i = 1}^p {\frac{{{2^{{r_i}}} - 1}}{{\log (1 + i)}}},
\end{equation}
where $p$ is the truncated position in a ranking list, \emph{Z} is a normalization constant to ensure that the NDCG score for the correct ranking is one, and $r_i$ is the similarity level of the \emph{i}-th database point in the ranking list. Directly optimizing such ranking criteria is intractable, which involves minimizing nonsmooth and multivariate ranking losses. One solution is to regard it as a problem of structured output learning and optimize the problem by structured SVM. But this framework is not suitable for deep learning model which is used to construct our hash functions. Next we will discuss a simple but effective scheme based on surrogate loss.

\begin{figure*}[t]
  \centering
  \subfigure[]{
  \begin{minipage}[b]{0.3\textwidth}
  \includegraphics[width=5.2cm]{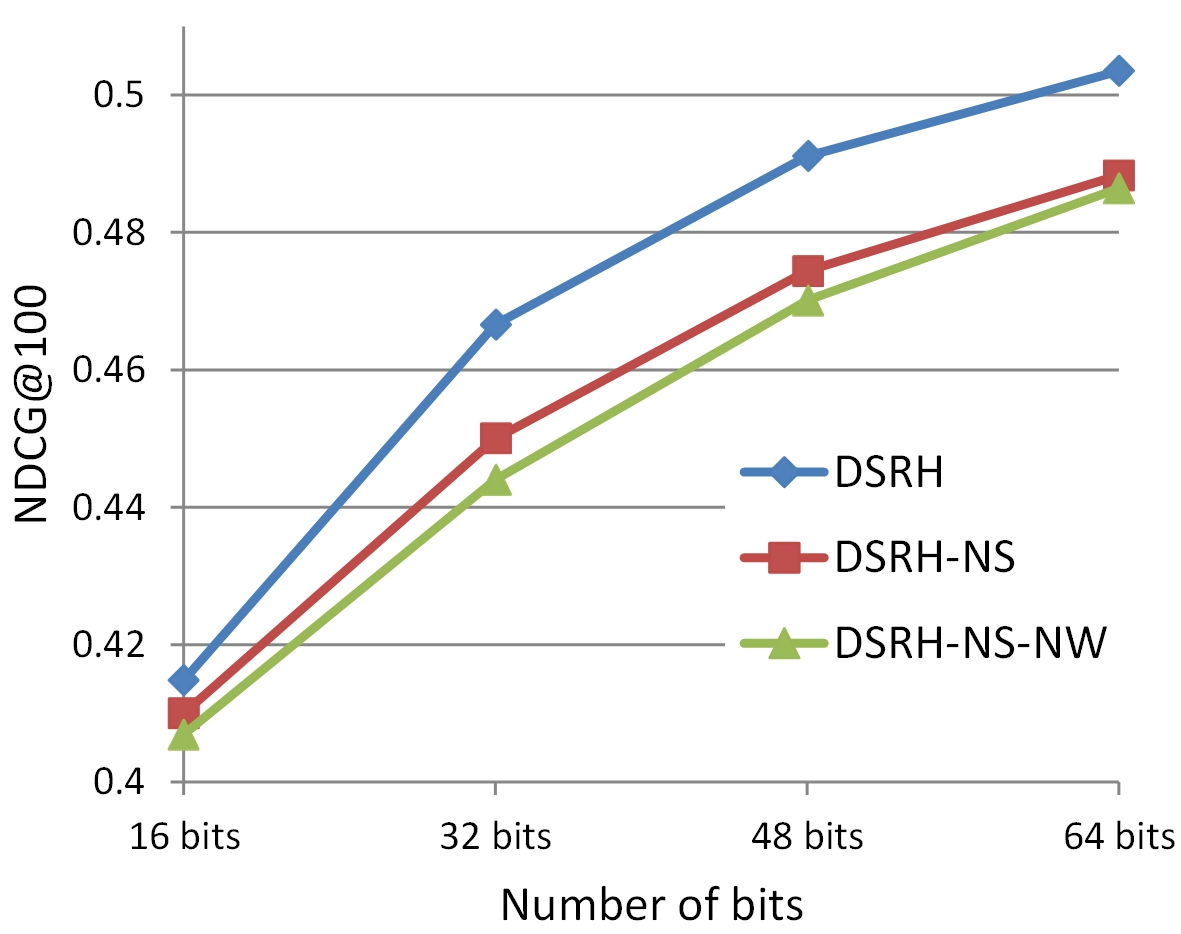}
  \end{minipage}

  \begin{minipage}[b]{0.3\textwidth}
  \includegraphics[width=5.2cm]{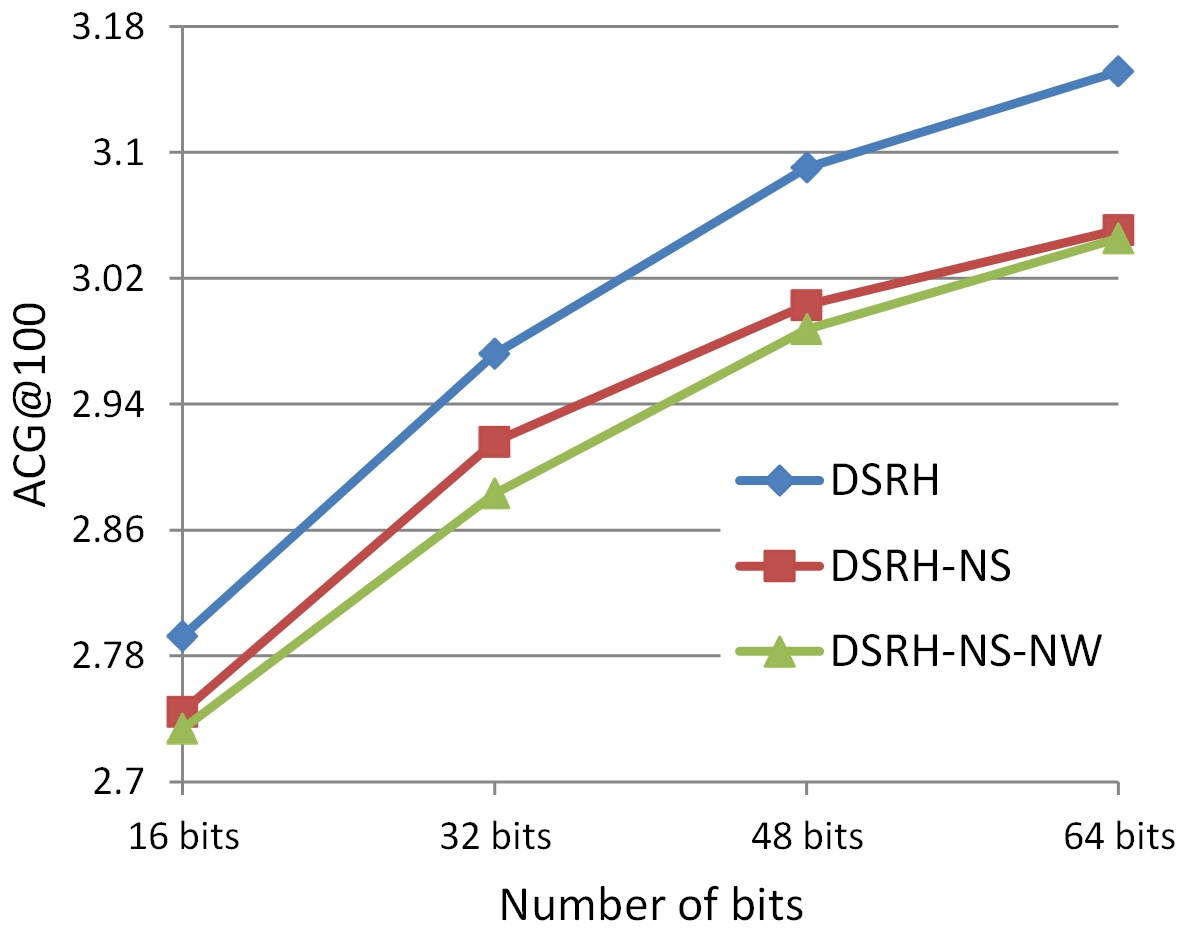}
  \end{minipage}

  \begin{minipage}[b]{0.3\textwidth}
  \includegraphics[width=5.2cm]{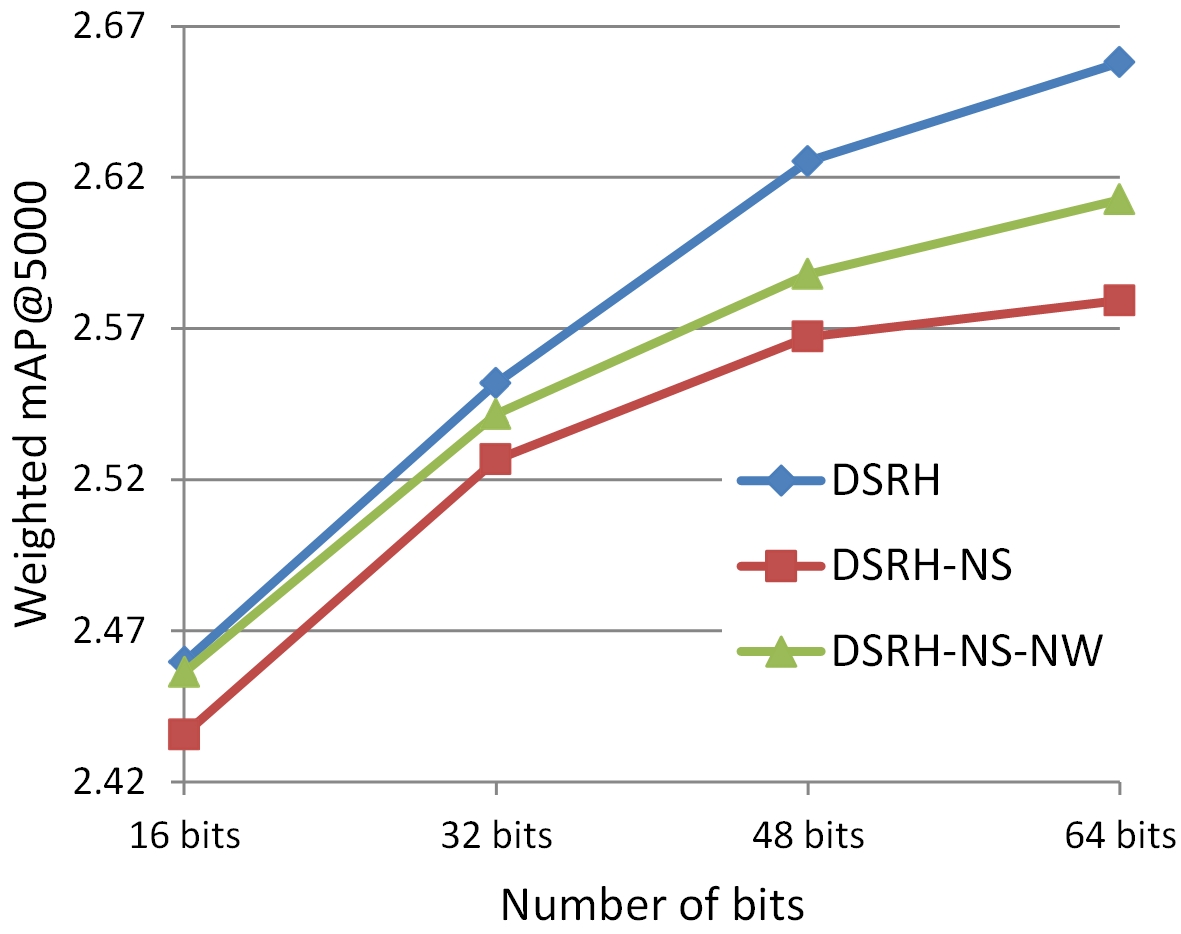}
  \end{minipage}
  }
  \subfigure[]{
  \begin{minipage}[b]{0.3\textwidth}
  \includegraphics[width=5.2cm]{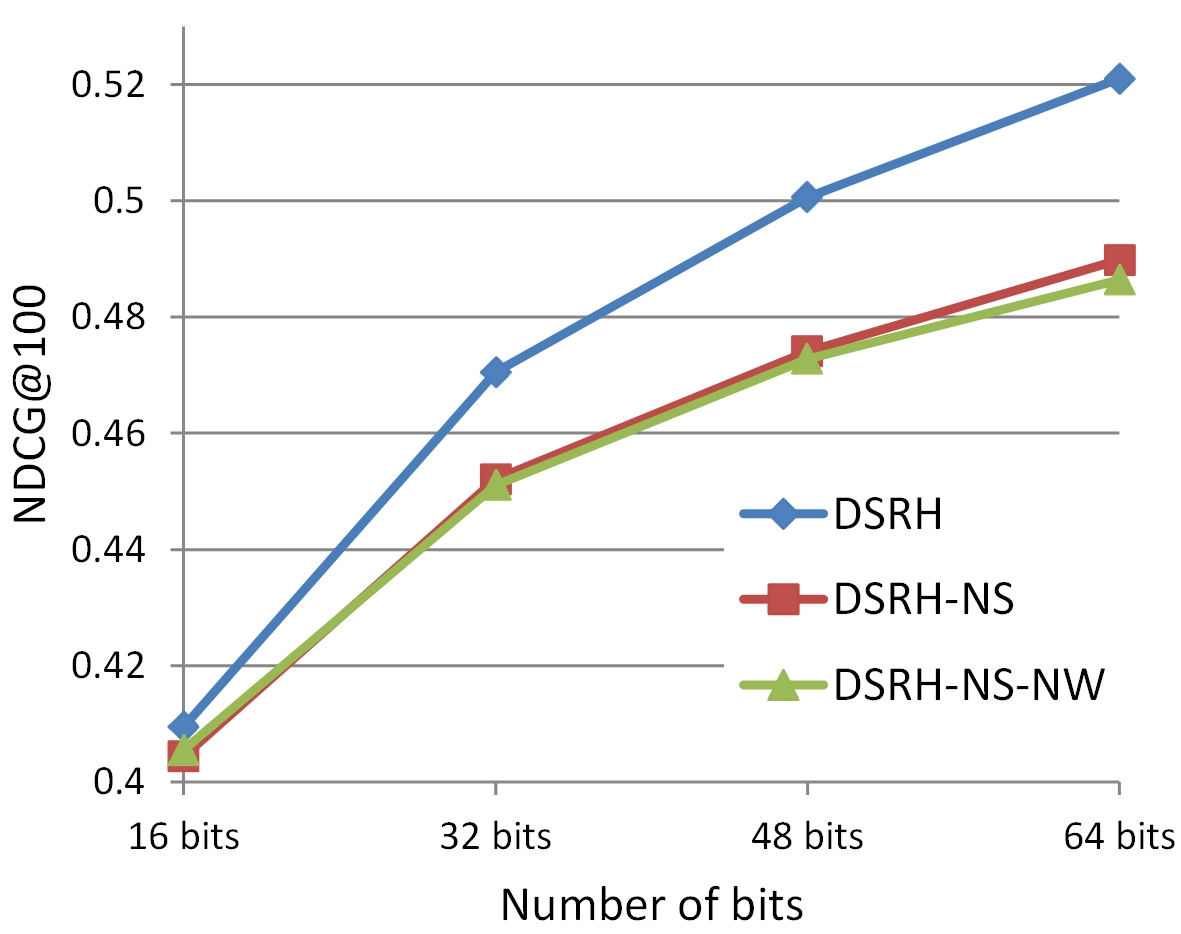}
  \end{minipage}

  \begin{minipage}[b]{0.3\textwidth}
  \includegraphics[width=5.2cm]{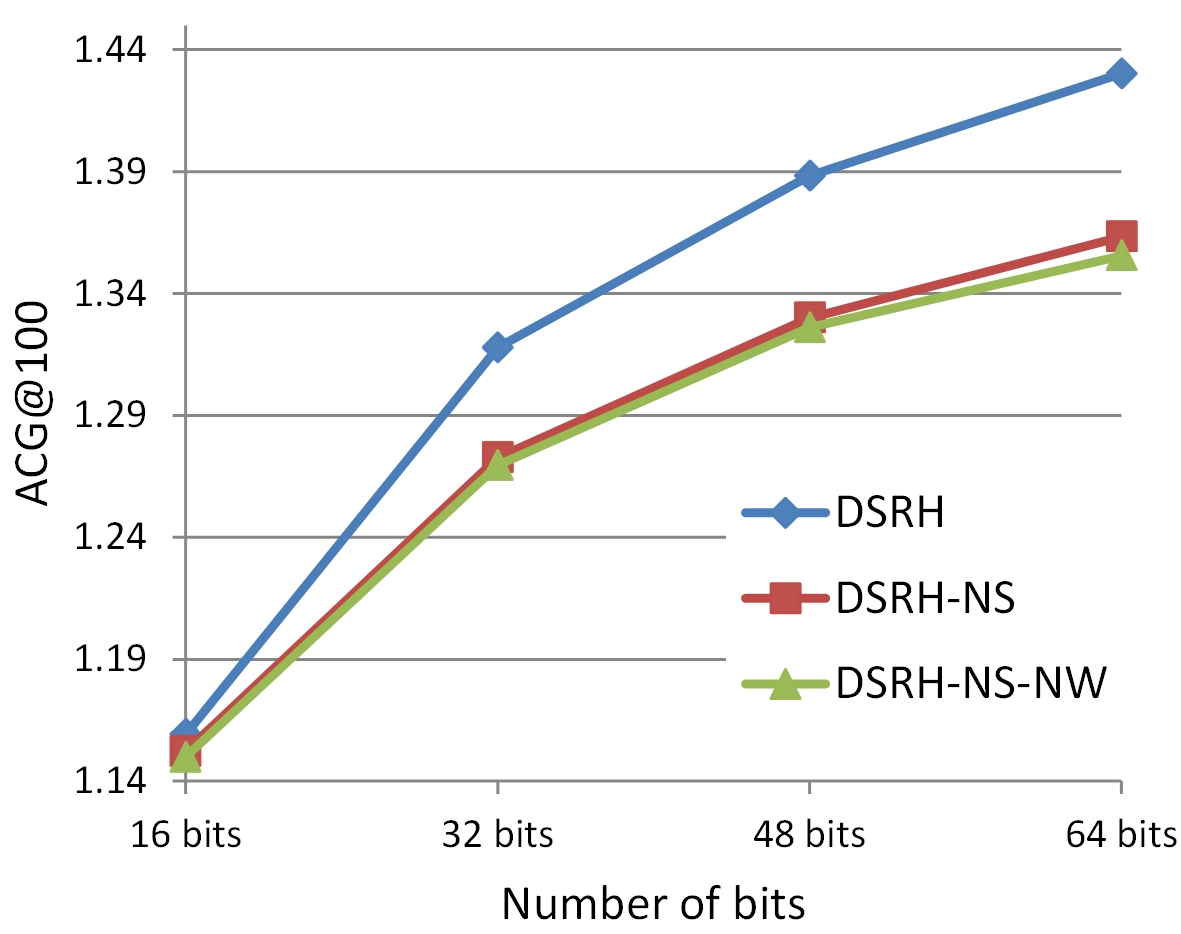}
  \end{minipage}

  \begin{minipage}[b]{0.3\textwidth}
  \includegraphics[width=5.2cm]{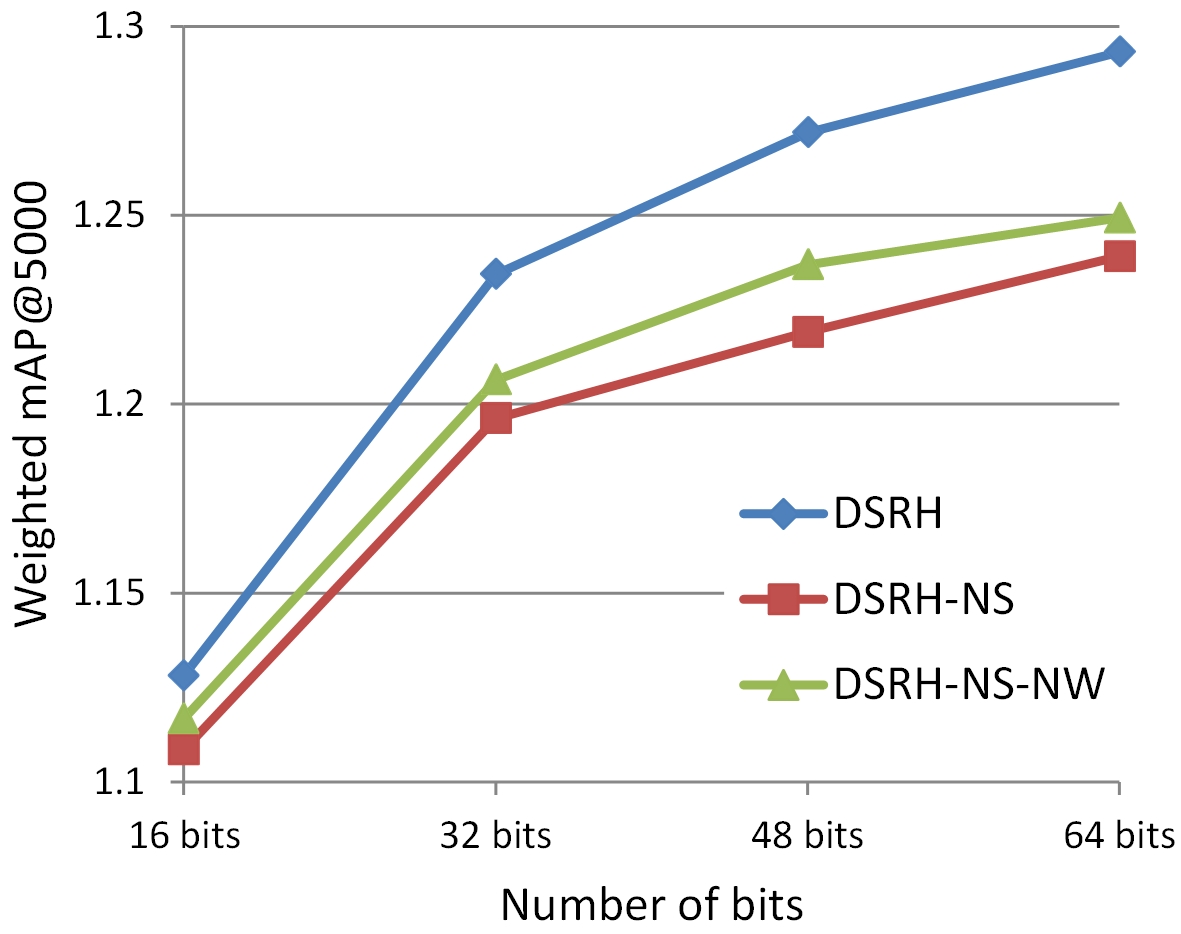}
  \end{minipage}
  }
  \caption{Ranking performance evaluations (NDCG, ACG and weighted mAP) of different components using various numbers of hash bits on two datasets: (a) MIRFLICKR-25K and (b) NUS-WIDE.} \label{fig_3}
\end{figure*}

\subsection{Optimization with Surrogate Loss}

To circumvent the problem of directly optimizing the ranking criteria, we try to use a surrogate loss as the risk that the learning procedure minimizes in practice. Given a query $\mathbf{q}$ and a ranking list $\{ {{\mathbf{x}}_i}\} _{i = 1}^M$ for $\mathbf{q}$, we can define a ranking loss on a set of triplets of hash codes as follows:
\begin{multline}\label{2}
  L({\mathbf{h}}({\mathbf{q}}),\{ {\mathbf{h}}({{\mathbf{x}}_i})\} _{i = 1}^M) = \\
  \sum\limits_{i = 1}^M {\sum\limits_{j:{r_j} < {r_i}} {{{\left[ {\delta {d_H}({\mathbf{h}}({\mathbf{q}}),{\mathbf{h}}({{\mathbf{x}}_i}),{\mathbf{h}}({{\mathbf{x}}_j})) + \rho } \right]}_ + }} },
\end{multline}
where $M$ is the length of the ranking list, ${[.]_ + } = \max (0,.)$, $\delta {d_H}({\mathbf{h}},{{\mathbf{h}}_1},{{\mathbf{h}}_2}) = {d_H}({\mathbf{h}},{{\mathbf{h}}_1}) - {d_H}({\mathbf{h}},{{\mathbf{h}}_2})$, $d_H(.,.)$ is the Hamming distance and $\rho$ is a margin parameter which controls the minimum margin between the distances of the two pairs. This surrogate loss is a convex upper bound on the pairwise disagreement which counts the number of incorrectly ranked triplets. This type of surrogate loss has been used in Ranking SVM \cite{R13} for leaning to rank where a score function for ranking is learned.

From the definition of NDCG, it can be observed that the top ranked items have a larger gain factor for the score, which better reflects the performance of the ranking models in practical image retrieval systems, because users usually pay most of their attentions to the results on the first few pages. Thus we wish that the ranking of these items could be predicted more accurately than others. However, (\ref{2}) treats all triplets equally, which is not desired. Inspired by \cite{R20}, we modify the ranking loss by adding adaptive weights related to the similarity levels of database points:
\begin{multline}\label{3}
  {L_\omega }({\mathbf{h}}({\mathbf{q}}),\{ {\mathbf{h}}({{\mathbf{x}}_i})\} _{i = 1}^M) = \\
  \sum\limits_{i = 1}^M {\sum\limits_{j:{r_j} < {r_i}} {\omega ({r_i},{r_j}){{\left[ {\delta {d_H}({\mathbf{h}}({\mathbf{q}}),{\mathbf{h}}({{\mathbf{x}}_i}),{\mathbf{h}}({{\mathbf{x}}_j})) + \rho } \right]}_ + }} }.
\end{multline}
According to NDCG, the weight $\omega$ can be given by:
\begin{equation}\label{4}
  \omega ({r_i},{r_j}) = \frac{{{2^{{r_i}}} - {2^{{r_j}}}}}{Z}
\end{equation}
where $Z$ is the normalization constant in (\ref{1}). The higher the relevance of ${\mathbf{x}}_i$ and $\mathbf{q}$ is than that of ${\mathbf{x}}_j$ and $\mathbf{q}$, the larger decline the NDCG score would suffer if ${\mathbf{x}}_i$ is ranked behind ${\mathbf{x}}_j$. And thus the larger weight should be assigned to this triplet. When $\omega ({r_i},{r_j}) \equiv 1$, it corresponds to (\ref{2}).

Given the dataset $\mathcal{D}$ as a training set, we wish to learn hash functions that optimize the rankings for all query points $\mathbf{q}$ from $\mathcal{D}$. Based on the surrogate loss (\ref{3}) and the hash function (\ref{5}), the objective function can be given by the empirical loss subject to some regularization:
\begin{multline}\label{6}
  \mathcal{F}({\mathbf{W}}) = \sum\limits_{{\mathbf{q}} \in \mathcal{D},\{ {{\mathbf{x}}_i}\} _{i = 1}^M \subset \mathcal{D}} {{L_\omega }({\mathbf{h}}({\mathbf{q}};{\mathbf{W}}),\{ {\mathbf{h}}({{\mathbf{x}}_i};{\mathbf{W}})\} _{i = 1}^M)} \\
  + \frac{\alpha }{2}\left\| {\mathop {\operatorname{mean} }\limits_{\mathbf{q}} ({\mathbf{h}}({\mathbf{q}};{\mathbf{W}}))} \right\|_2^2 + \frac{\beta }{2}\left\| {\mathbf{W}} \right\|_2^2.
\end{multline}
Similar to \cite{R10}, the second term is the balance penalty which is used to encourage each bit averaged over the training data to be mean-zero and to make sure more stable convergence of the learning procedure. And the third term is the $L_2$ weight decay which penalizes large weights \cite{R22}. Due to the discontinuous sign function in (\ref{5}), the optimization of (\ref{6}) is difficult. To address this issue, we relax $h({\mathbf{x}};{\mathbf{w}})$ to:
\begin{equation}\label{7}
  h({\mathbf{x}};{\mathbf{w}}) = 2\sigma ({{\mathbf{w}}^{\rm T}}[{f_a}({\mathbf{x}});{f_b}({\mathbf{x}})]) - 1,
\end{equation}
where $\sigma (t) = 1/(1 + \exp ( - t))$ is the logistic function. In order to facilitate the gradient computation, we rewrite the hamming distance as the form of inner product:
\begin{equation}\label{8}
  {d_H}({\mathbf{h}}({\mathbf{q}};{\mathbf{W}}),{\mathbf{h}}({\mathbf{x}};{\mathbf{W}})) = \frac{{K - {\mathbf{h}}{{({\mathbf{q}};{\mathbf{W}})}^{\rm T}}{\mathbf{h}}({\mathbf{x}};{\mathbf{W}})}}{2}
\end{equation}
where $K$ is the number of hash bits.

Stochastic gradient descent is used to minimize the objective function. It can be observed that the loss function (\ref{3}) is actually a summation of a sequence of weighted triplet losses. For any one triplet $(\mathbf{q},{\mathbf{x}}_i,{\mathbf{x}}_j)$, if
\begin{equation*}
  \frac{1}{2}({\mathbf{h}}{({\mathbf{q}};{\mathbf{W}})^{\rm T}}{\mathbf{h}}({{\mathbf{x}}_j};{\mathbf{W}}) - {\mathbf{h}}{({\mathbf{q}};{\mathbf{W}})^{\rm T}}{\mathbf{h}}({{\mathbf{x}}_i};{\mathbf{W}})) + \rho  > 0,
\end{equation*}
the derivatives of (\ref{6}) with respect to hash code vectors are given by:
\begin{multline}\label{9}
  \frac{{\partial {\mathcal{F} }}}{{\partial {\mathbf{h}}({\mathbf{q}};{\mathbf{W}})}} = \frac{\alpha }{{{N_q}}}\mathop {\operatorname{mean} }\limits_{\mathbf{q}} ({\mathbf{h}}({\mathbf{q}};{\mathbf{W}}))\\
  + \frac{1}{2}\omega ({r_i},{r_j})({\mathbf{h}}({{\mathbf{x}}_j};{\mathbf{W}}) - {\mathbf{h}}({{\mathbf{x}}_i};{\mathbf{W}}))
\end{multline}
\begin{gather}
  \frac{{\partial {\mathcal{F} }}}{{\partial {\mathbf{h}}({{\mathbf{x}}_i};{\mathbf{W}})}} =  - \frac{1}{2}\omega ({r_i},{r_j}){\mathbf{h}}({\mathbf{q}};{\mathbf{W}}), \label{10} \\
  \frac{{\partial {\mathcal{F} }}}{{\partial {\mathbf{h}}({{\mathbf{x}}_j};{\mathbf{W}})}} = \frac{1}{2}\omega ({r_i},{r_j}){\mathbf{h}}({\mathbf{q}};{\mathbf{W}}). \label{11}
\end{gather}
where the mean value is computed over one mini-batch and $N_q$ is the size of a mini-batch. These derivative values can be fed into the underlying CNN via the back-propagation algorithm to update the parameters of each layer.

\begin{figure*}[t]
  \centering
  \subfigure[]{
  \begin{minipage}[b]{0.3\textwidth}
  \includegraphics[width=5.2cm]{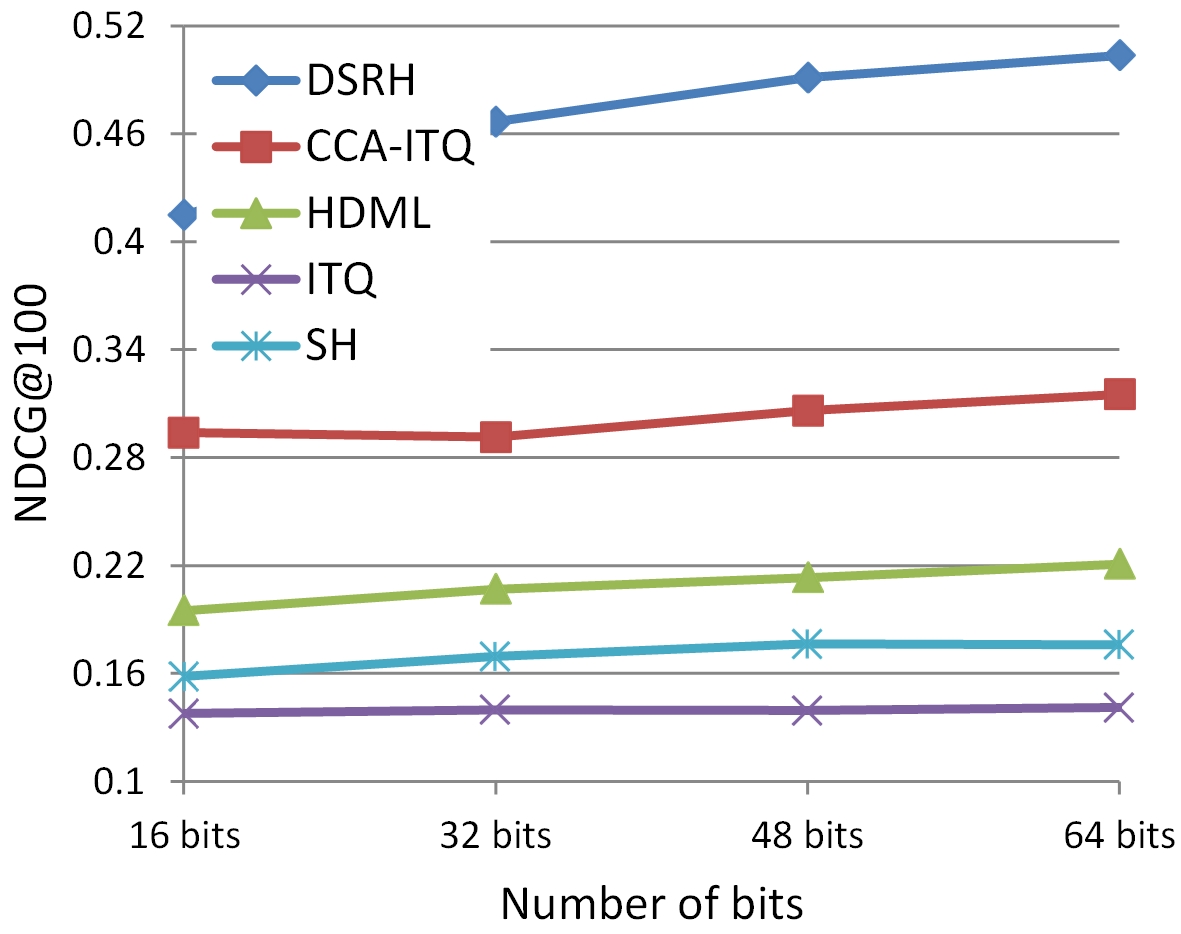}
  \end{minipage}

  \begin{minipage}[b]{0.3\textwidth}
  \includegraphics[width=5.2cm]{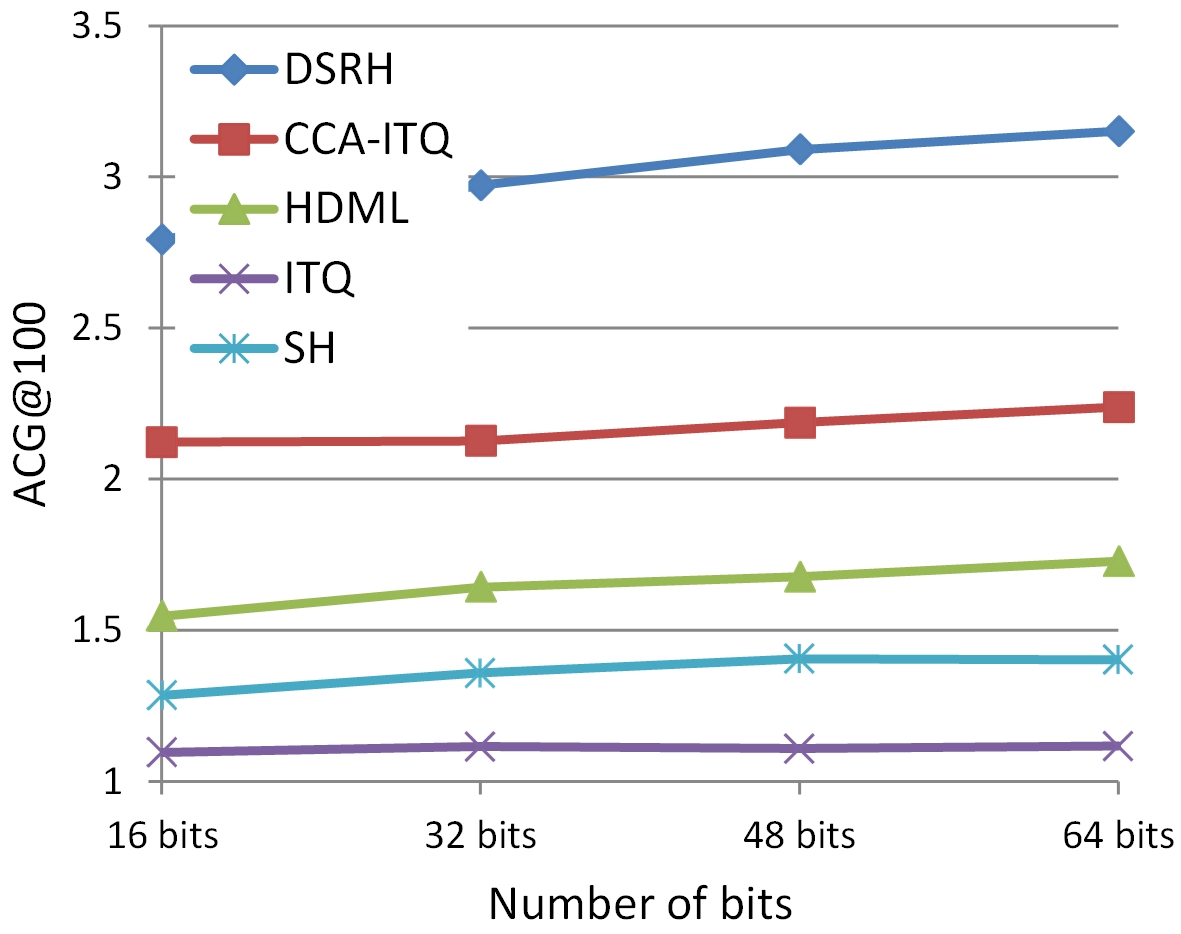}
  \end{minipage}

  \begin{minipage}[b]{0.3\textwidth}
  \includegraphics[width=5.2cm]{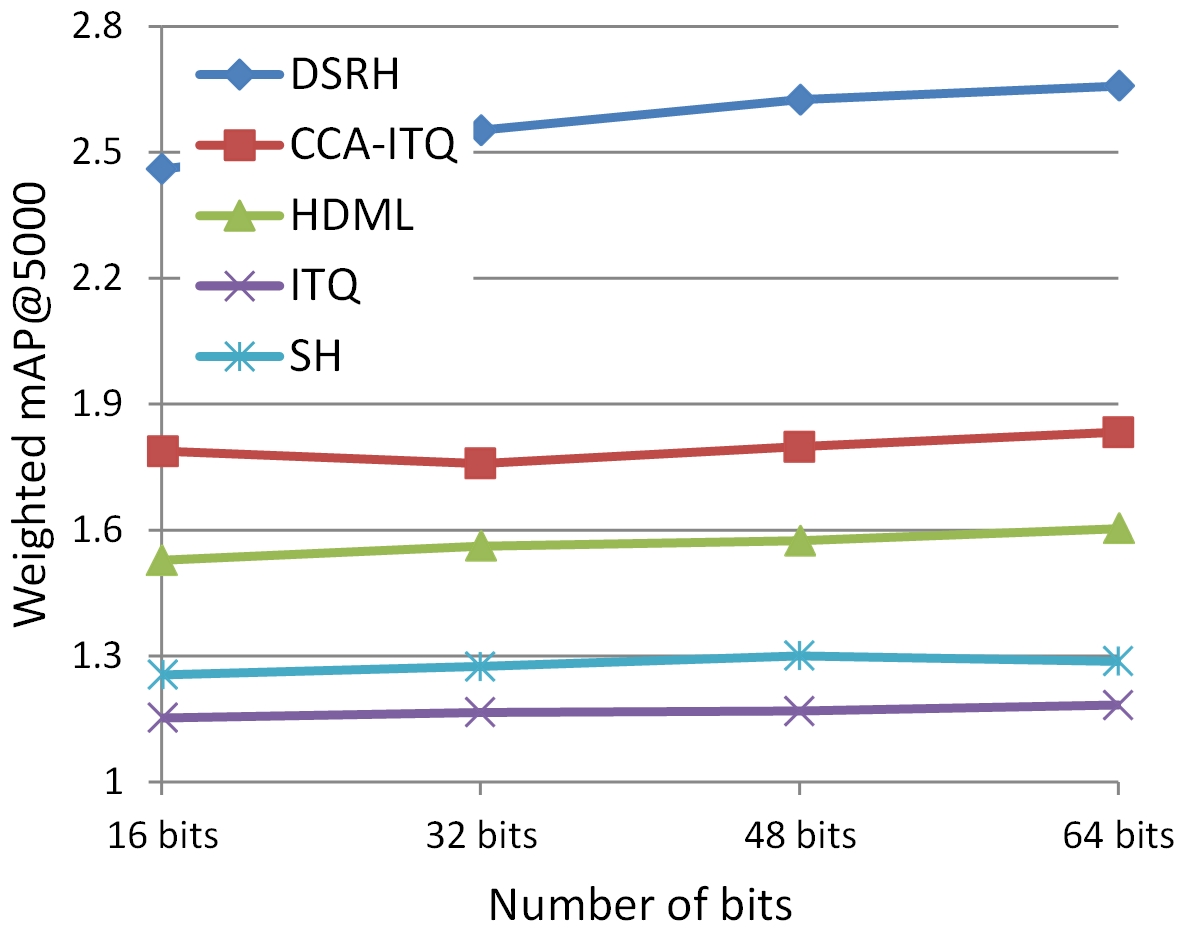}
  \end{minipage}
  }
  \subfigure[]{
  \begin{minipage}[b]{0.3\textwidth}
  \includegraphics[width=5.2cm]{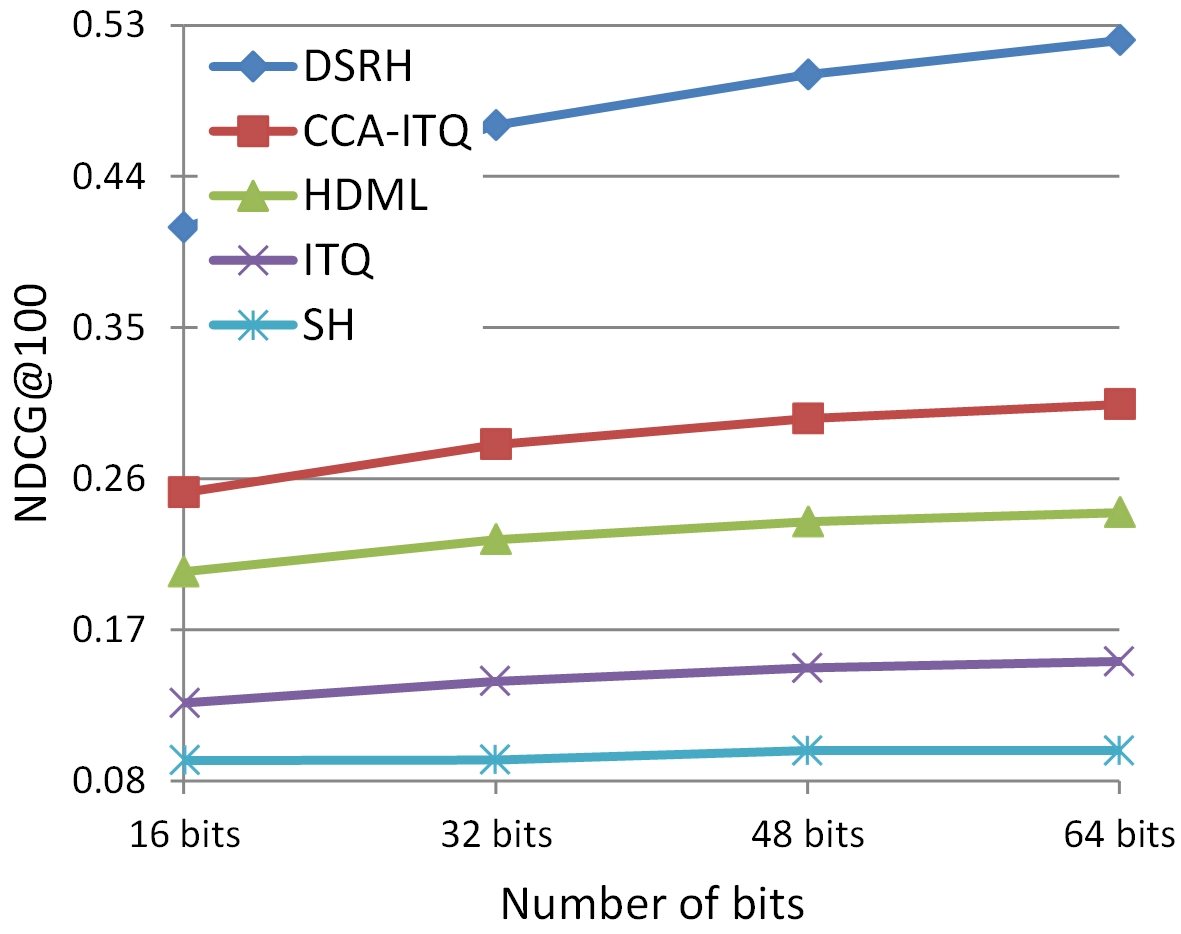}
  \end{minipage}

  \begin{minipage}[b]{0.3\textwidth}
  \includegraphics[width=5.2cm]{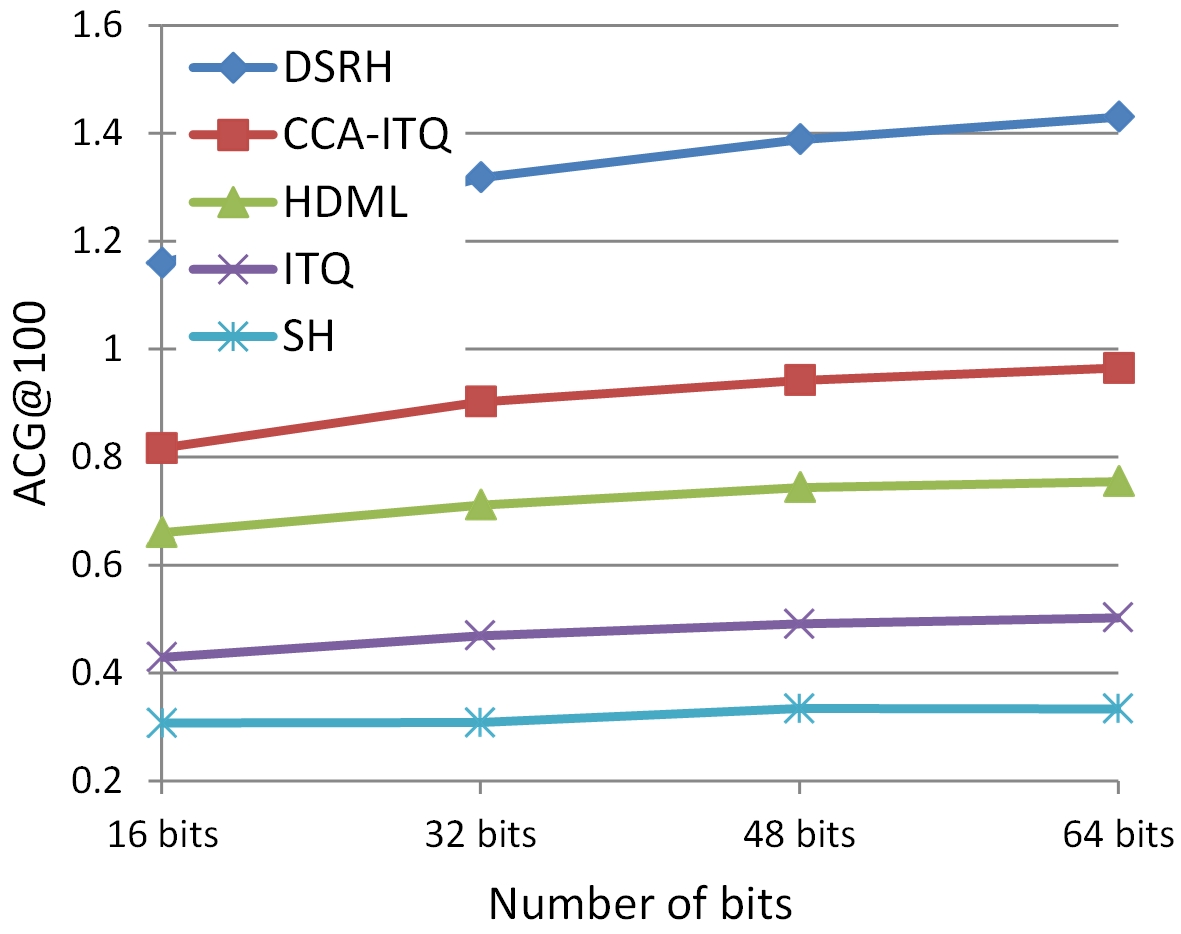}
  \end{minipage}

  \begin{minipage}[b]{0.3\textwidth}
  \includegraphics[width=5.2cm]{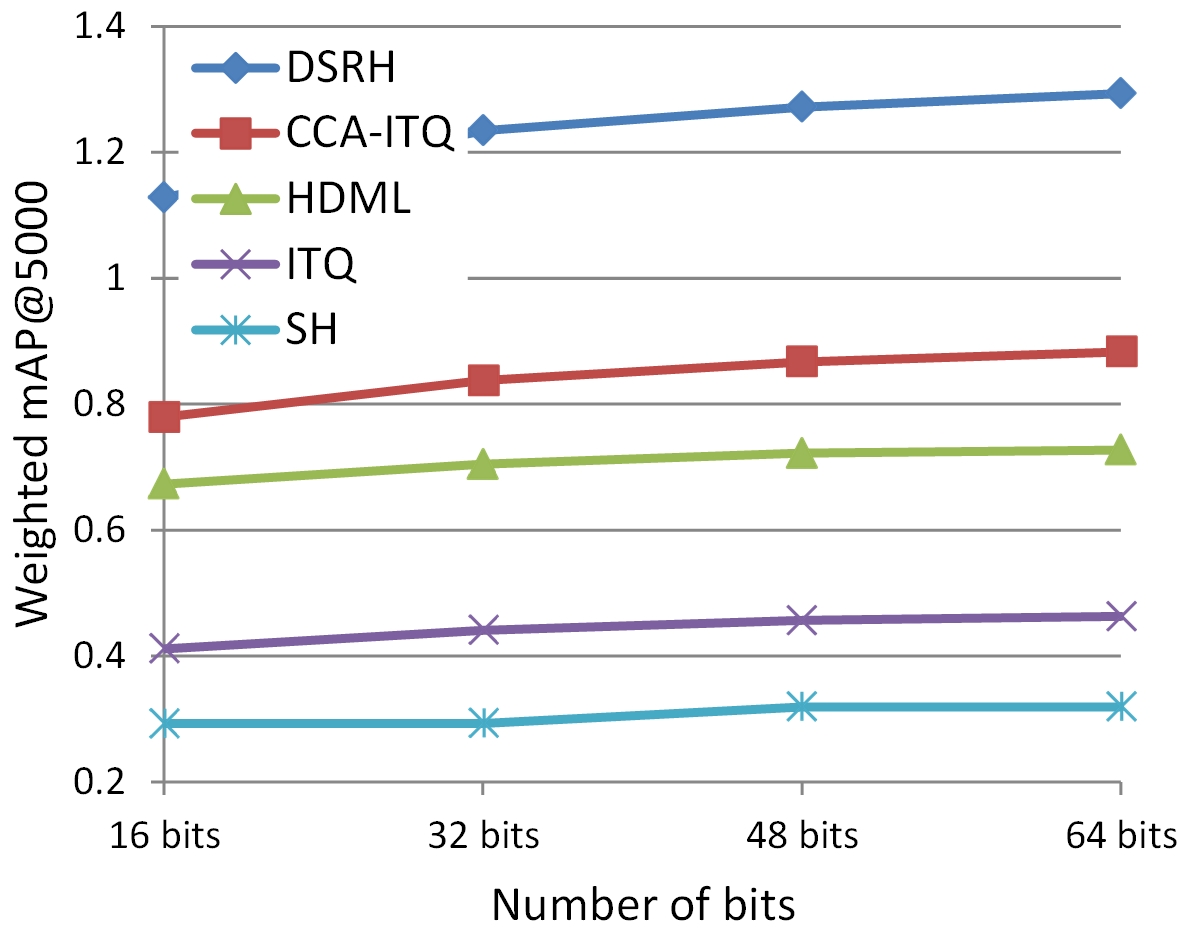}
  \end{minipage}
  }
  \caption{Comparison of ranking performance of our DSRH and other hashing methods based on hand-crafted features on two datasets: (a) MIRFLICKR-25K and (b) NUS-WIDE.} \label{fig_4}
\end{figure*}

\section{Experiments}

We test the proposed hashing method on two multi-label benchmark datasets, i.e., MIRFLICKR-25K \cite{R24} and NUS-WIDE \cite{R18}. We present quantitative evaluations in terms of ranking measures and compare our method with unsupervised methods: iterative quantization (ITQ) \cite{R5}, spectral hashing (SH) \cite{R2}, and supervised methods using multi-label and ranking information respectively: CCA-ITQ \cite{R5}, hamming distance metric learning (HDML) \cite{R10}.

We set the mini-batch size for gradient descent to 128, and impose dropout with keeping probability 0.5 on the fully connected layers to avoid overfitting. The regularization parameter $\alpha$ and $\beta$ in the objective function (\ref{6}) are set to 1 and $5e^{-4}$ respectively. The length of the ground-truth ranking list used for training is set to 3, which can be created by taking one item sharing all the labels with a query, one item without any common label and one item having at least one common label. For all compared methods, we use the best settings reported in their literatures.

The ImageNet ILSVRC-2012 dataset \cite{R25} is utilized to pre-train the CNN model by optimizing multinomial logistic regression objective function in the image classification task. This dataset contains about 1.2 million training images and 50,000 validation images, roughly 1000 images in each of 1000 categories.  We use the pre-trained parameters of convolutional layers and fully-connected layers to initialize the CNN part of hash functions in our method. In order to ensure fairness, we apply the features from the pre-trained CNN model to the compared hashing methods as well. Furthermore, the compared methods are carried out using the features that are learned through fine-tuning the CNN model on the multi-label datasets in our retrieval task.

\subsection{Datasets}

The MIRFLICKR-25K dataset \cite{R24} consists of 25,000 images collected from the social photography website Flickr. All images are annotated for 24 semantic concepts including various scenes and objects categories such as sky, night, food and tree. Moreover, 14 of these concepts are used for a stricter labeling, i.e., an image is annotated with a concept again only if the concept is salient. Thus we have total 38 semantic labels where each image may belong to several labels. 2000 images are randomly selected as testing queries and the remaining images are used as the database for training and retrieval. Following \cite{R26}, a 3857-dimensional feature vector for each image is extracted by concatenating Pyramid Histogram of Words (PHOW) features, Gist and MPEG-7 descriptors, which will be used for the compared methods.

The NUS-WIDE dataset \cite{R18} is a relatively larger image dataset containing 269,648 images annotated with 81 concepts. It is also from Flickr, but more challenging than MIRFLICKR-25K due to more images and more diverse semantic concepts. Since the website of the dataset does not provide raw image data but the URLs of images, for learning deep feature representations we download these images ourselves from the web. However, we only collected 226,265 images as some of those URLs have been invalid now. We randomly sample 5000 images for testing queries and the rest is used for training and retrieval. The dataset includes six types of low-level features extracted from these images: bag of words based on SIFT feature, color histogram, color correlogram, edge direction histogram, wavelet texture and block-wise color moments. We concatenate them all and get a 1134-dimensional feature representation for each image.

\begin{figure*}[t]
  \centering
  \subfigure[]{
  \begin{minipage}[b]{0.3\textwidth}
  \includegraphics[width=5.2cm]{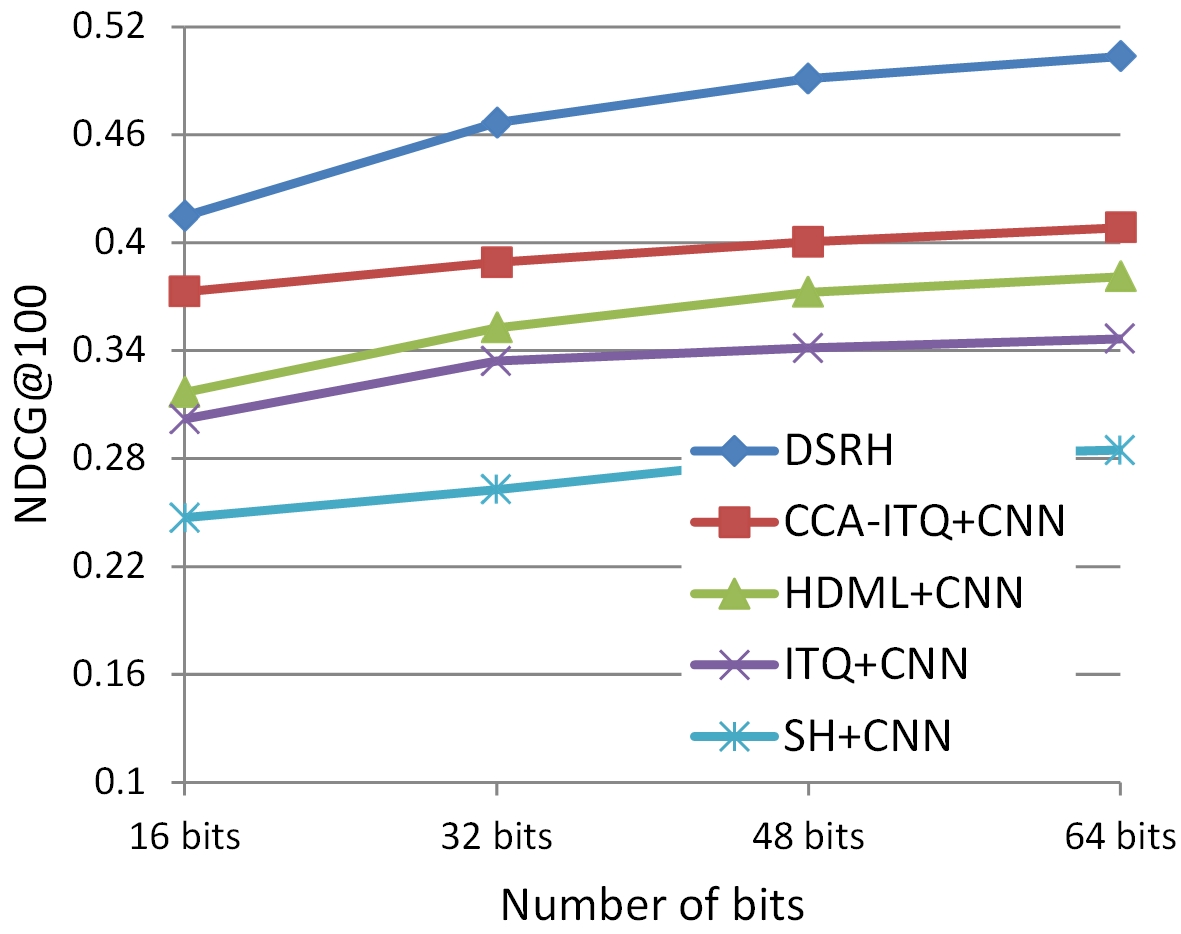}
  \end{minipage}

  \begin{minipage}[b]{0.3\textwidth}
  \includegraphics[width=5.2cm]{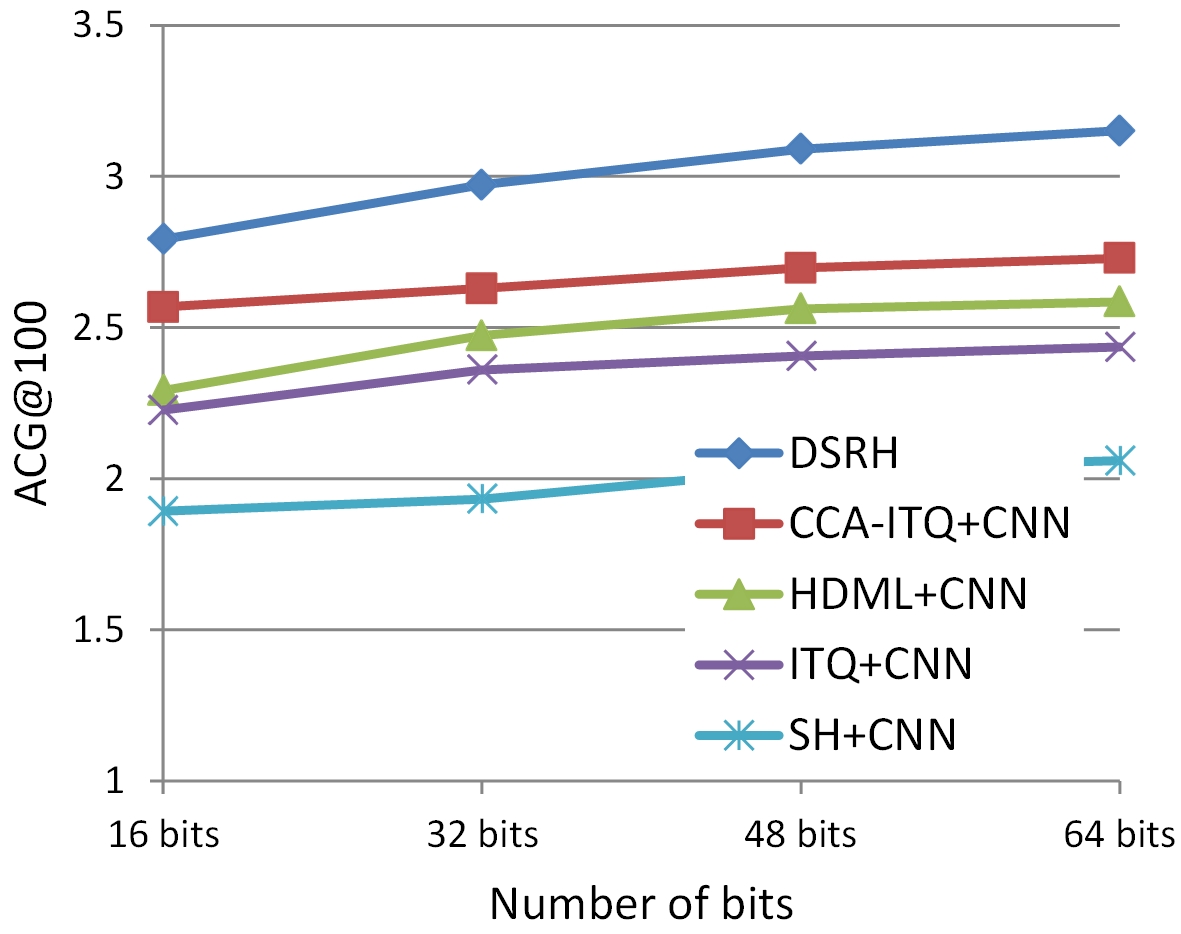}
  \end{minipage}

  \begin{minipage}[b]{0.3\textwidth}
  \includegraphics[width=5.2cm]{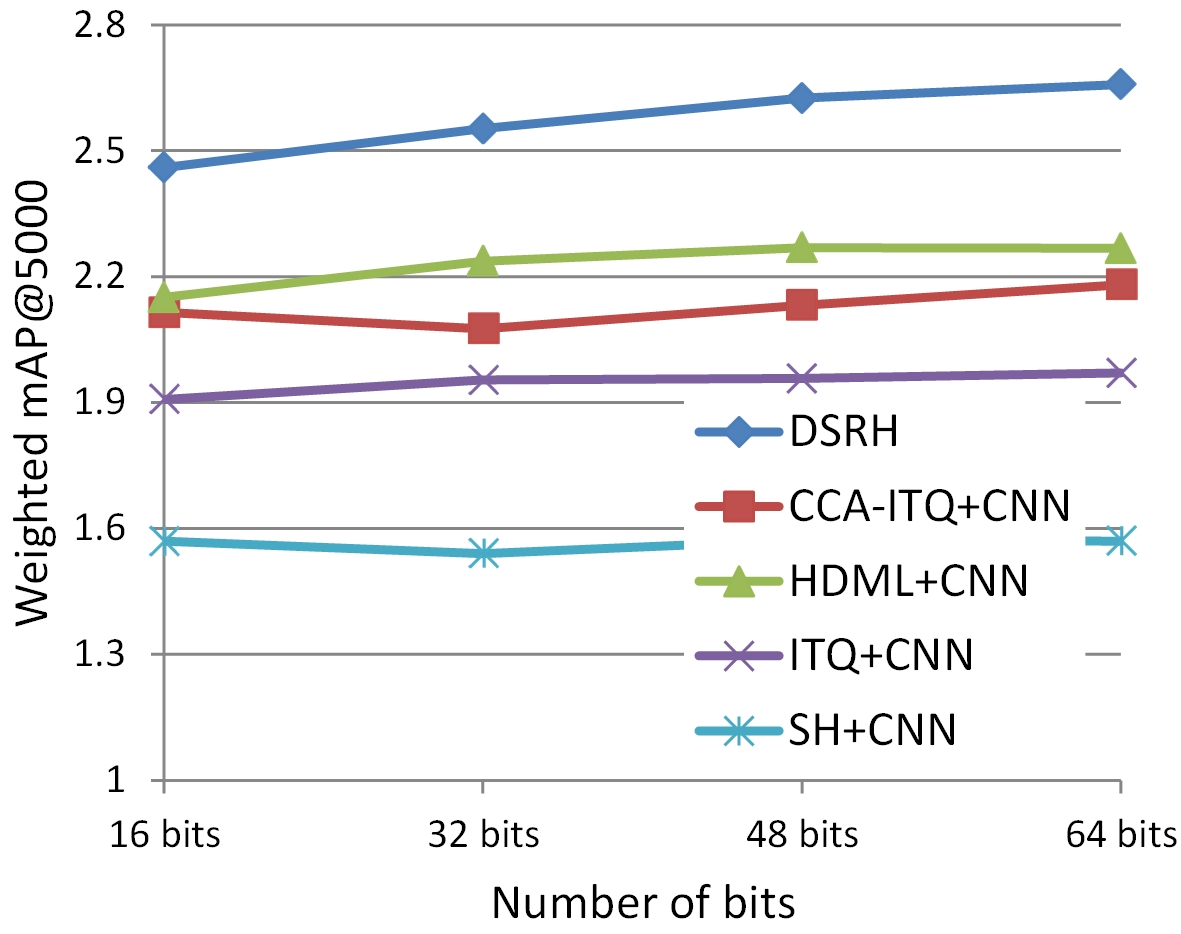}
  \end{minipage}
  }
  \subfigure[]{
  \begin{minipage}[b]{0.3\textwidth}
  \includegraphics[width=5.2cm]{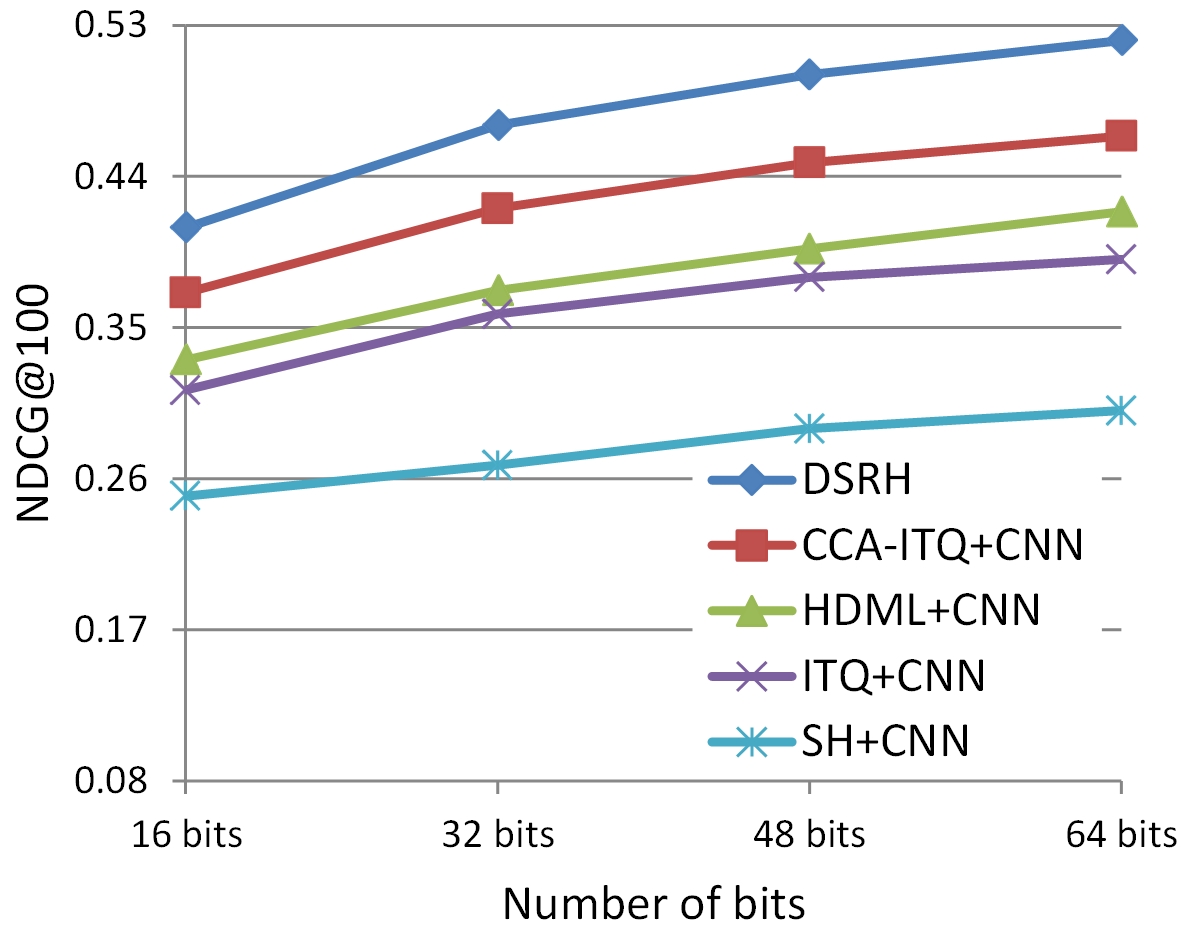}
  \end{minipage}

  \begin{minipage}[b]{0.3\textwidth}
  \includegraphics[width=5.2cm]{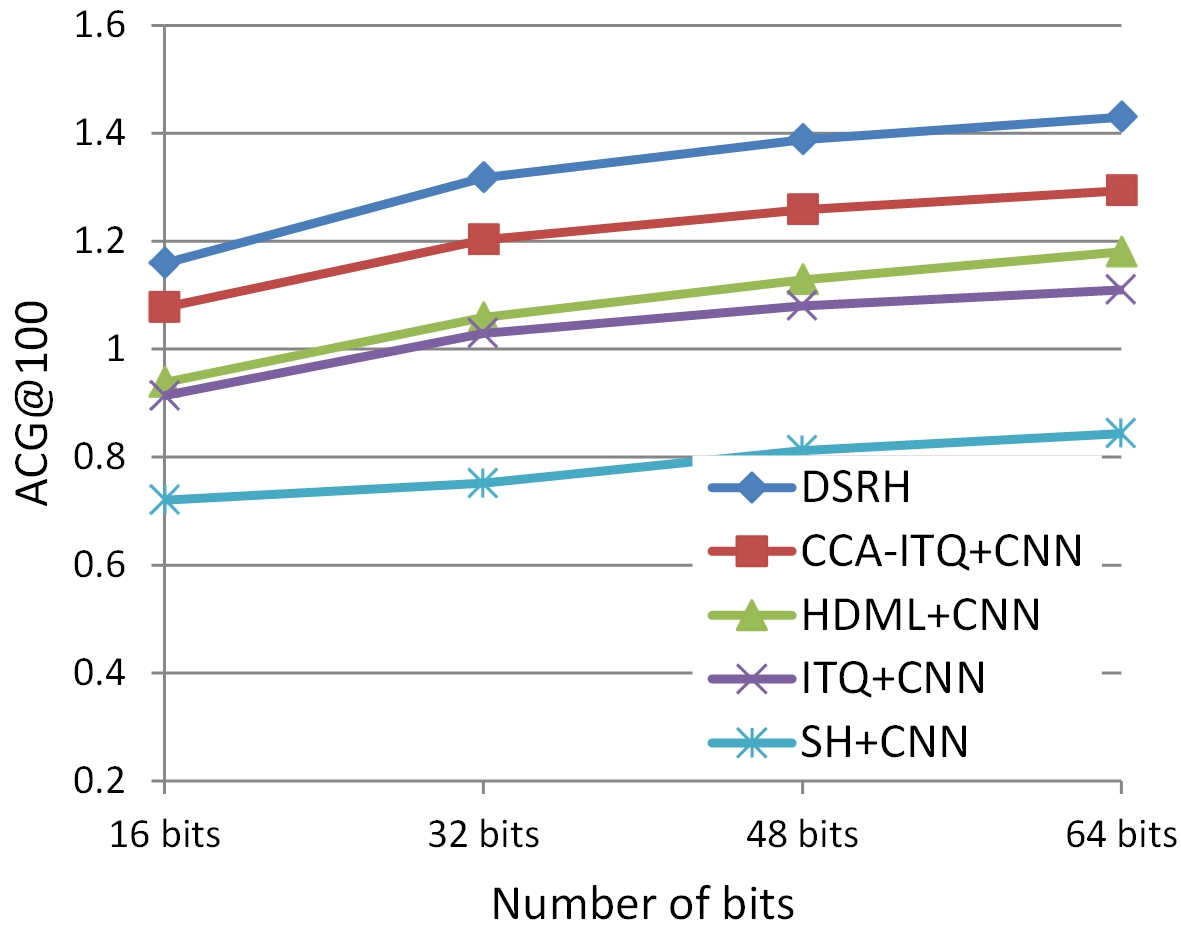}
  \end{minipage}

  \begin{minipage}[b]{0.3\textwidth}
  \includegraphics[width=5.2cm]{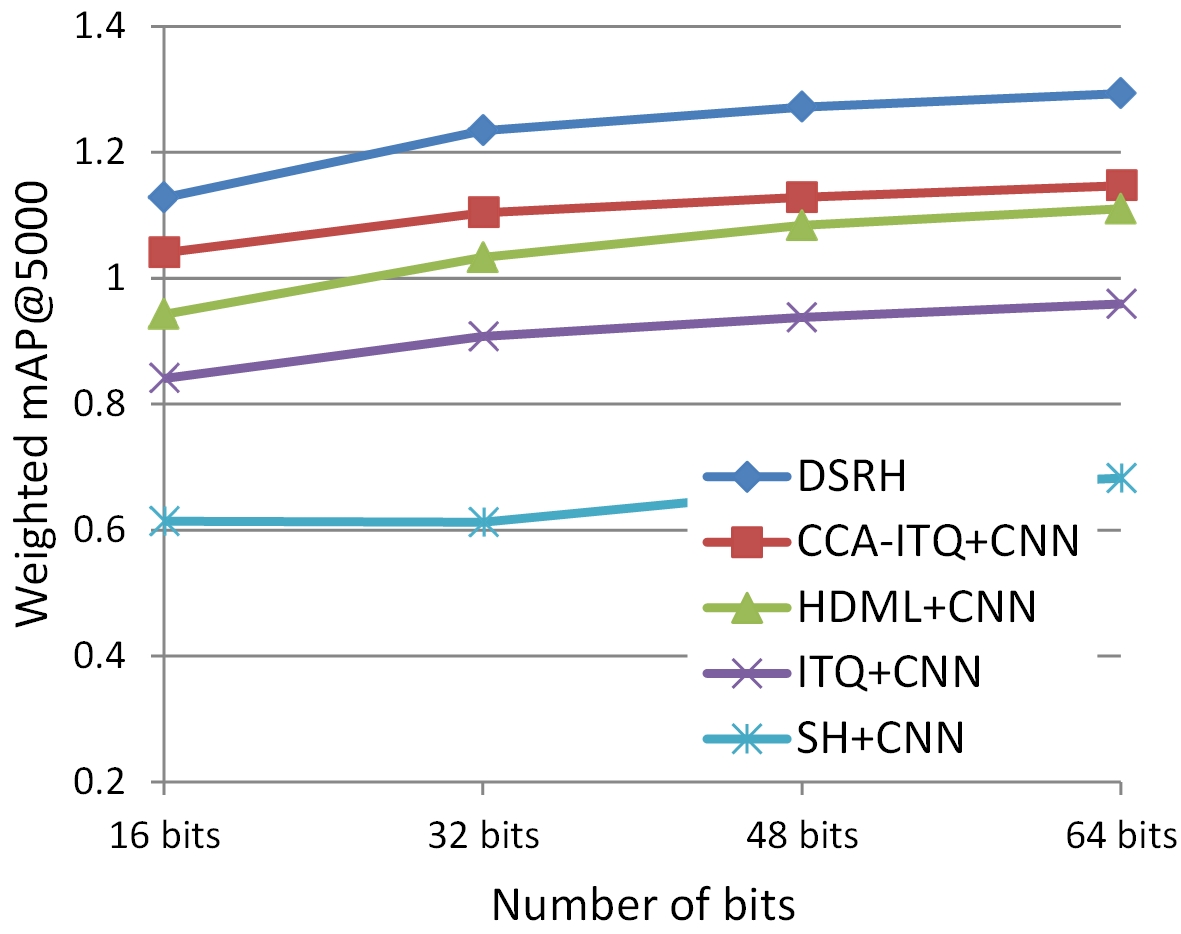}
  \end{minipage}
  }
  \caption{Comparison of ranking performance of our DSRH and other hashing methods based on activation features of pre-trained CNN on two datasets: (a) MIRFLICKR-25K and (b) NUS-WIDE.} \label{fig_5}
\end{figure*}

\subsection{Evaluation Criteria}

In our experiments, Normalized Discounted Cumulative Gain (NDCG) \cite{R21}, Average Cumulative Gain (ACG) \cite{R21} and weighted mean Average Precision (mAP) are used to measure the ranking quality of retrieved database points. As mentioned before, NDCG defined in (\ref{1}) evaluates the ranking of data points by penalizing errors in higher ranked items more strongly.

ACG is calculated by taking the average of the similarity levels of data points within top-$p$ positions:
\begin{equation}\label{12}
  ACG@p = \frac{1}{p}\sum\limits_{n = 1}^p {{r_i}},
\end{equation}
where $r_i$ is the similarity level of the data point on the \emph{i}-th position of a ranking. ACG actually is equivalent to the precision weighted by the similarity level of each data point.

mAP is the mean of average precision for each query. Since the similarity level $r$ can be bigger than one, we compute a weighted mAP using ACG:
\begin{gather}\label{13}
  {mAP_w} = \frac{1}{Q}\sum\limits_{q = 1}^Q {A{P_w}(q)}, \\
  {AP_w} = \frac{{\sum\nolimits_{p = 1}^M {\Pi ({r_p} > 0)ACG@p} }}{{{M_{r > 0}}}},
\end{gather}
where $\Pi (.) \in \left\{ {0,1} \right\}$ is an indicator function and $M_{r > 0}$ is the number of relevant data points.

\subsection{Evaluation of Different Components}

To analyze the effectiveness of several important components in the proposed DSRH, we remove the connection between the hash layer and the skipping layer and set $\omega ({r_i},{r_j}) \equiv 1$ in (\ref{3}) to evaluate their influence on the final performance. These two models are called DSRH-NS and DSRH-NS-NW. Here NS denotes no skipping layer and NW denotes no adaptive weight. Fig. \ref{fig_3} shows the results in the ranking measures.

We can see that using the surrogate loss with adaptive weights can improve the ranking quality of top-100 relevant items in terms of NDCG and ACG at the expense of the averaged ranking performance because it assigns larger weights to more relevant database points and weakens the effect of the less relevant ones. Connecting the first fully-connected layer to the hash layer can also improve the performance because more information biased toward visual appearance can be utilized which may be important for capturing multilevel semantic similarity.

\begin{figure*}[t]
  \centering
  \subfigure[]{
  \begin{minipage}[b]{0.3\textwidth}
  \includegraphics[width=5.2cm]{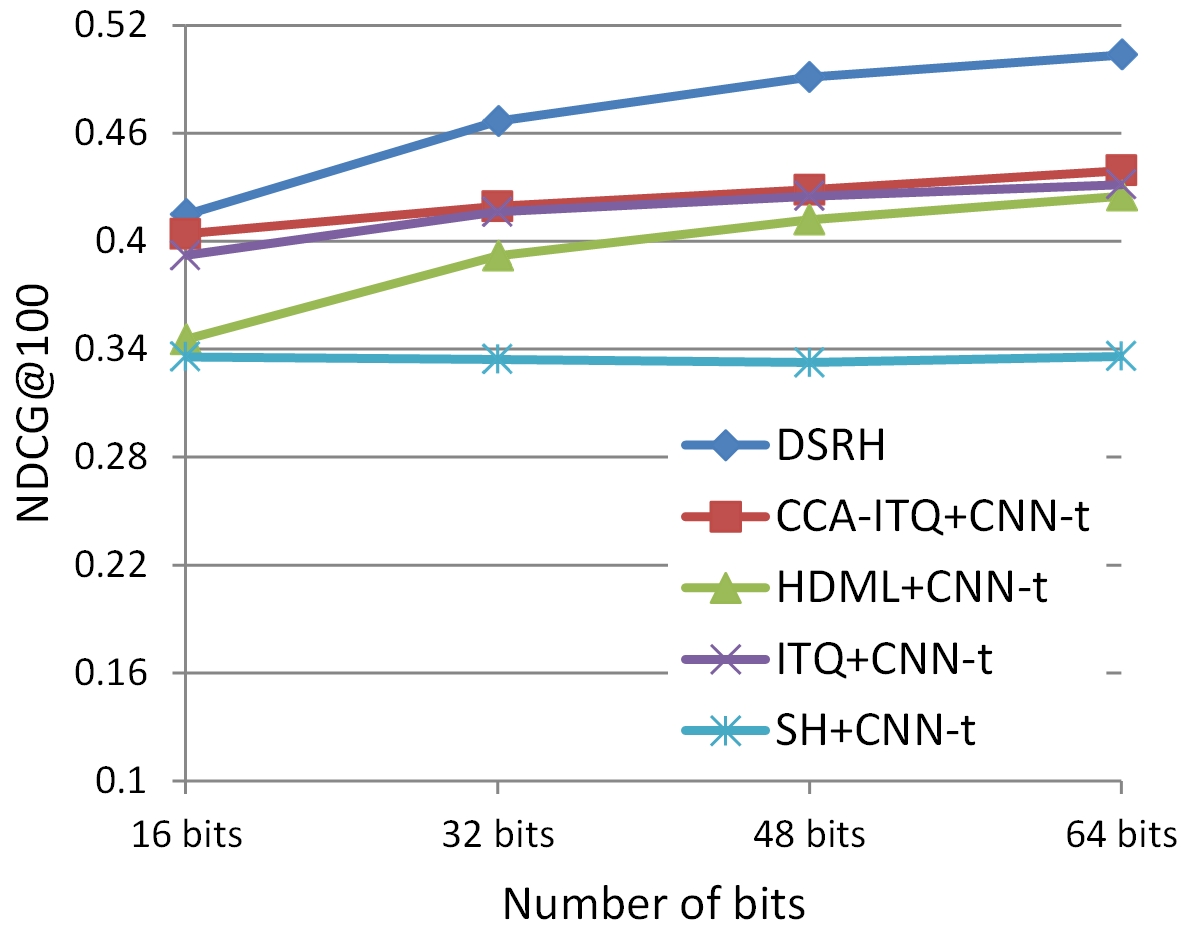}
  \end{minipage}

  \begin{minipage}[b]{0.3\textwidth}
  \includegraphics[width=5.2cm]{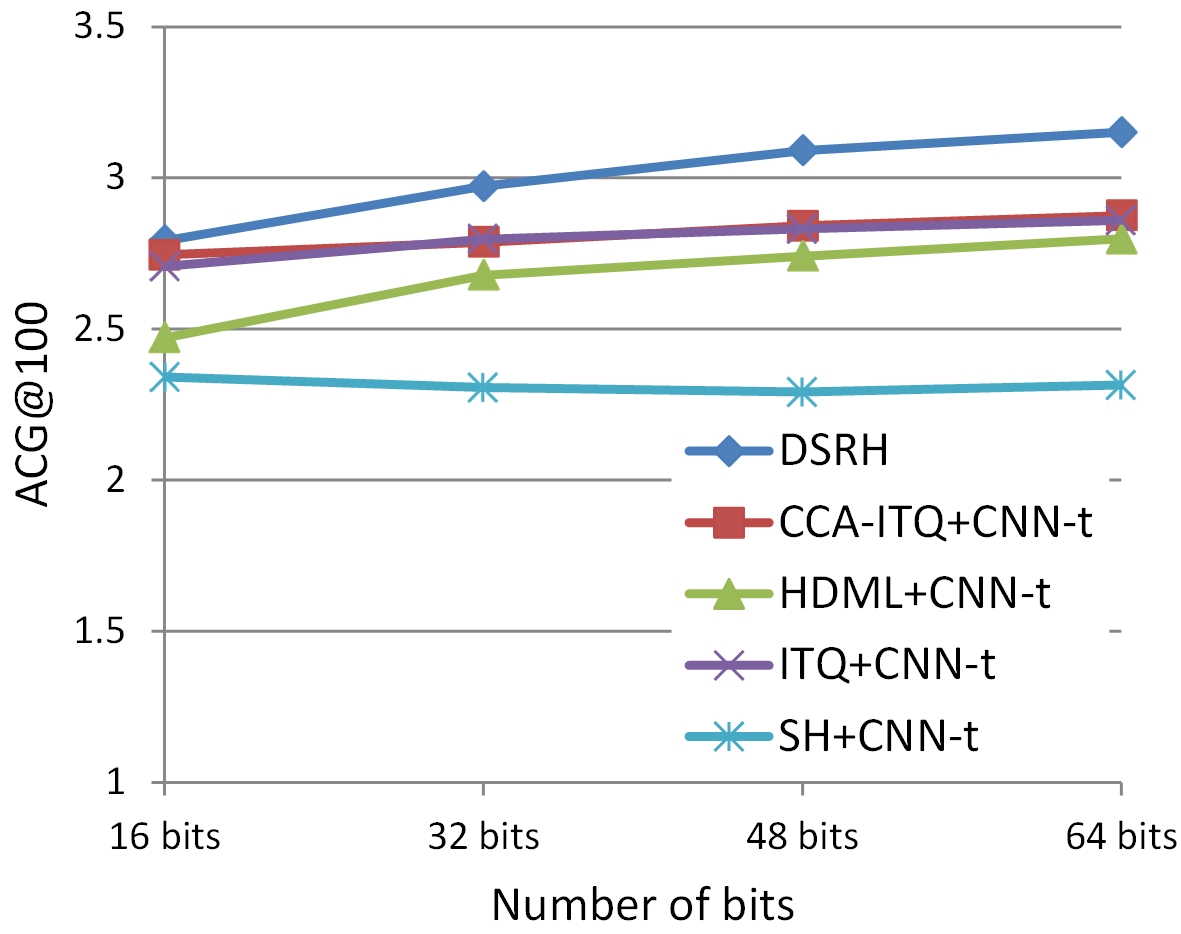}
  \end{minipage}

  \begin{minipage}[b]{0.3\textwidth}
  \includegraphics[width=5.2cm]{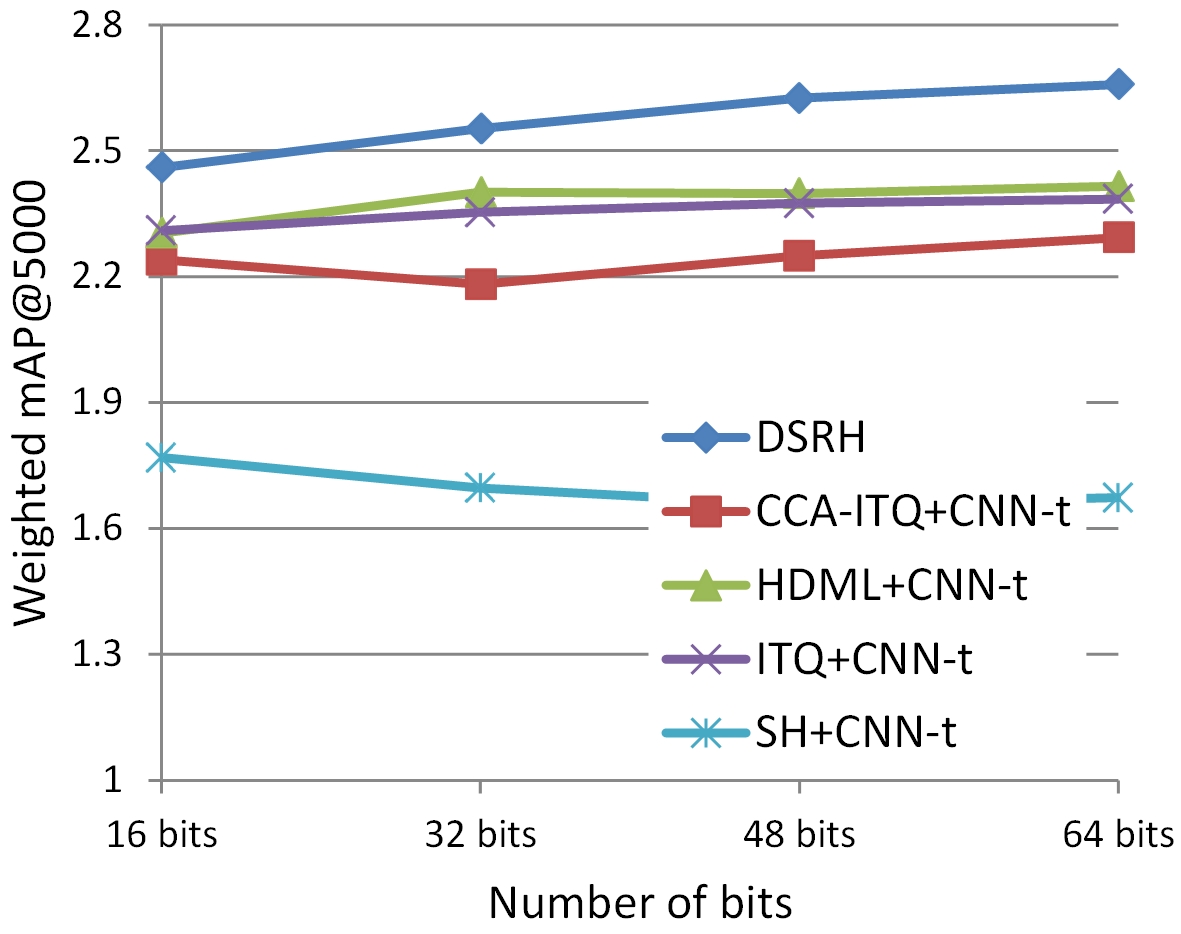}
  \end{minipage}
  }
  \subfigure[]{
  \begin{minipage}[b]{0.3\textwidth}
  \includegraphics[width=5.2cm]{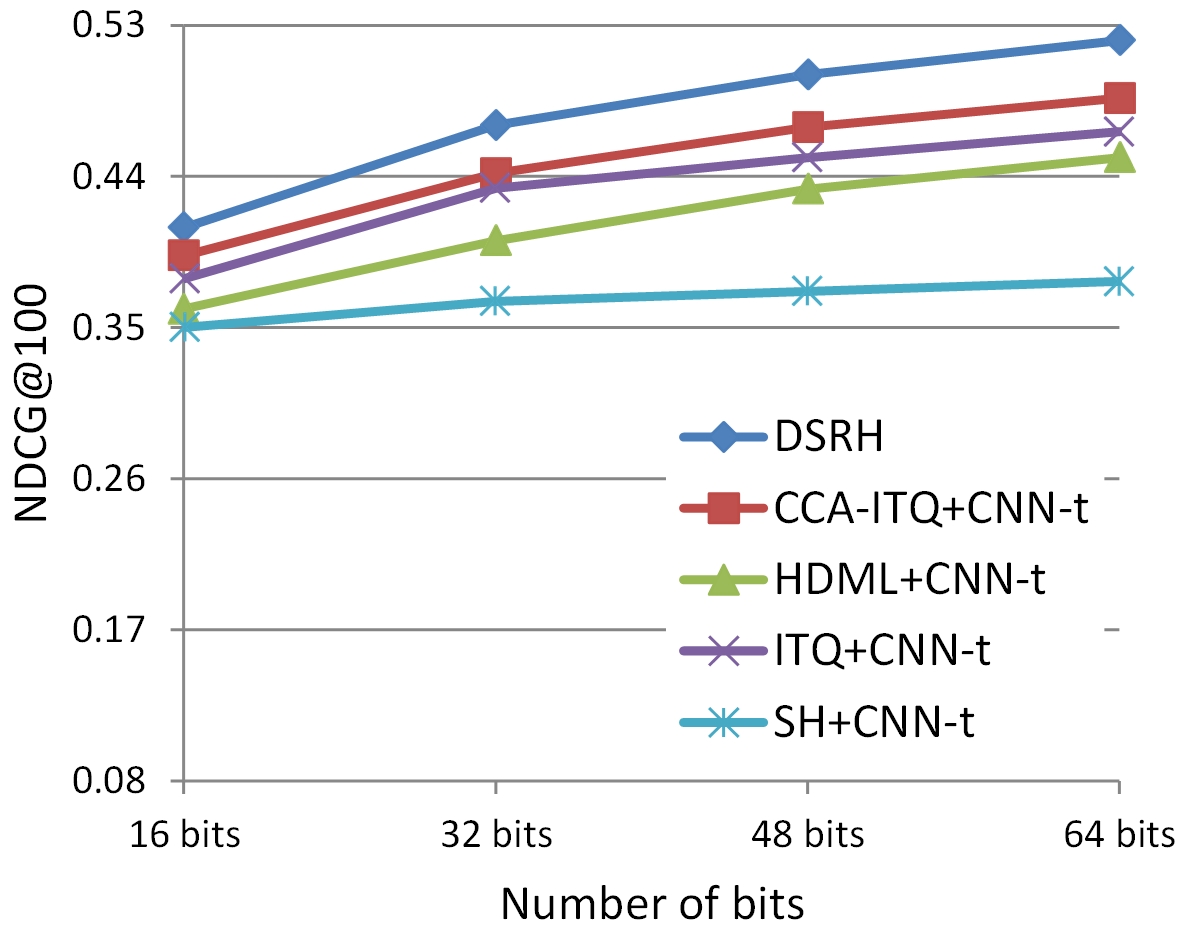}
  \end{minipage}

  \begin{minipage}[b]{0.3\textwidth}
  \includegraphics[width=5.2cm]{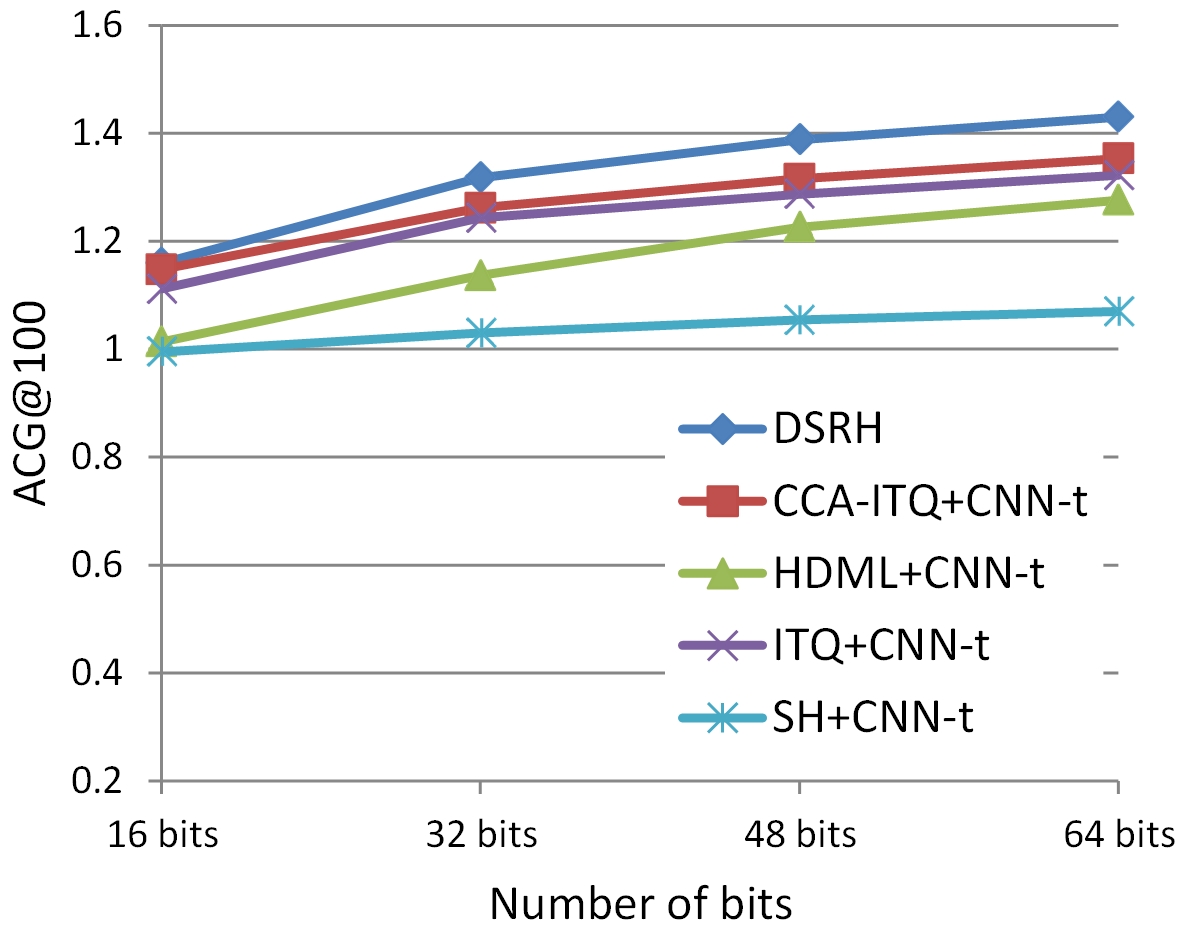}
  \end{minipage}

  \begin{minipage}[b]{0.3\textwidth}
  \includegraphics[width=5.2cm]{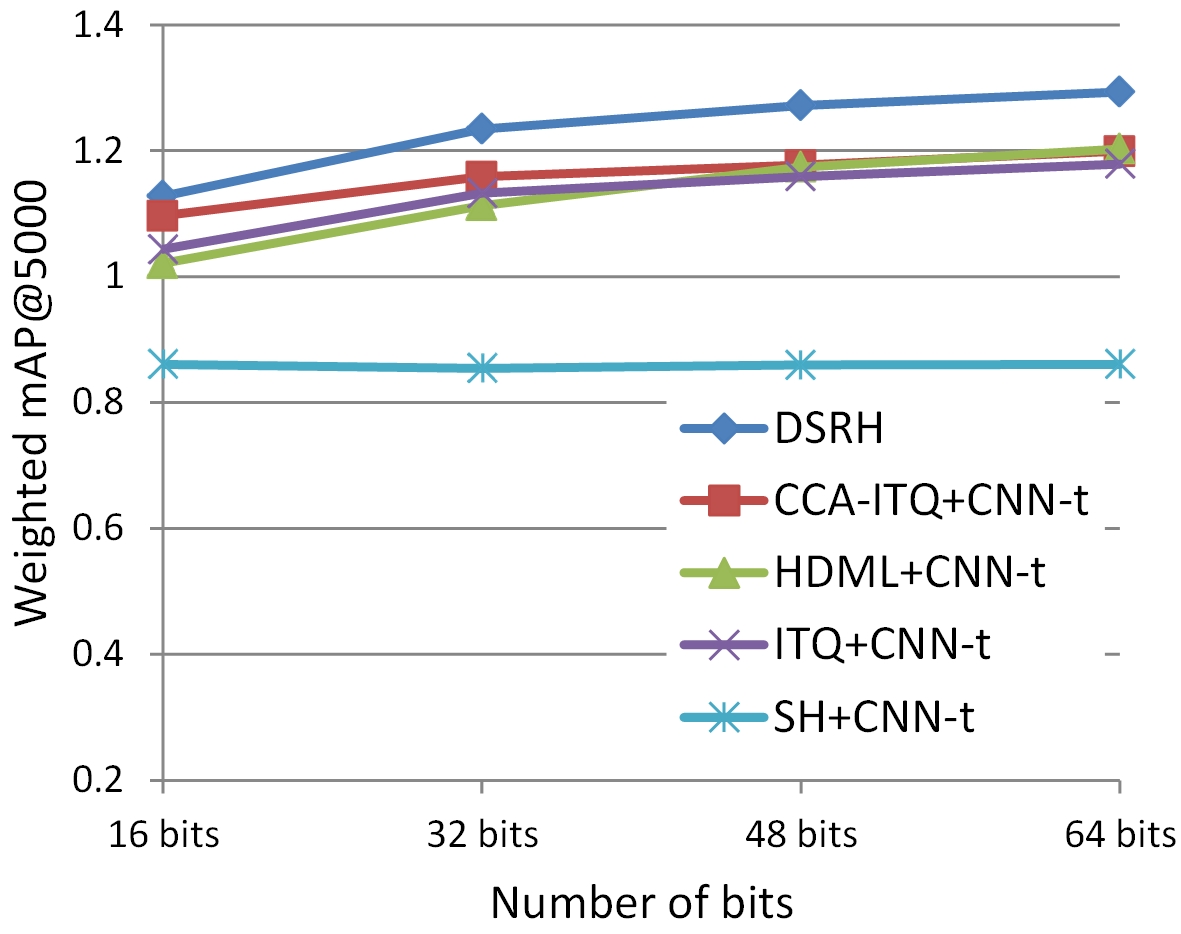}
  \end{minipage}
  }
  \caption{Comparison of ranking performance of our DSRH and other hashing methods based on activation features of fine-tuned CNN on two datasets: (a) MIRFLICKR-25K and (b) NUS-WIDE.} \label{fig_6}
\end{figure*}

\subsection{Method Comparison}

We also compare the proposed DSRH with other hash methods based on hand-crafted features. Fig. \ref{fig_4} illustrates the scores of NDCG, ACG and weighted mAP of these methods using various numbers of bits. We can see that the performance of our method is significantly better than other methods based on hand-crafted features in all cases. By using the CNN model to construct hash functions, our method have higher learning capability and is able to exploit more semantic information than the hashing methods trained on hand-crafted features which are usually extracted by unsupervised and shallow models.

We further evaluate the compared hashing methods on the features obtained from the activation of the last hidden layer of the CNN model pre-trained on the ImageNet dataset. This activation feature can be seen as a generic visual feature and has been used in object recognition, domain adaption and scene recognition \cite{R27}. The results of NDCG, ACG and weighted mAP are shown in Fig. \ref{fig_5}. Although the activation features boost the ranking performance of the compared methods by a large margin, our method still has better performance because we construct the deep hash function to jointly learning feature representations and hash codes, which can utilize semantic supervision information to obtain features more fitted to the retrieval datasets. It is more effective than learning hash codes from the features learned in advance.

To further verify the superiority of our method, we use the activation features fine-tuned on the retrieval datasets, i.e., MIRFLICKR-25K and NUS-WIDE, to evaluate the compared methods as well. Specifically, we retrain the CNN model for multi-label image classification by using cross-entropy cost function. We report the results in Fig. \ref{fig_6}. It can be seen that our method still achieves the best ranking performance, which shows that the multilevel semantic ranking supervision can make hash function learning better preserve the semantic structure of multi-label images. CCA-ITQ also uses multi-label information, but dose not explicitly learn with the ranking. HDML performs even worse than the unsupervised ITQ because it only considers binary similarity relationship which harms the semantic structure in the fine-tuned features. Note that using the features fine-tuned by multi-label supervision, the unsupervised ITQ performs almost as well as CCA-ITQ which is its supervised version, even better in terms of weighted mAP.

Similar to the skipping layer in the hash function of DSRH, we also attempt to concatenate the activations of the last two hidden layers of the CNN model as feature representations and apply them to the compared hashing methods. However, the performance of the compared methods trained using these features become even worse. It further validates the effectiveness of the structure of our hash function which has a tight coupling with CNN.

%

\section{Conclusion}

In this paper we have proposed to employ multilevel semantic ranking supervision to learn deep hash functions based on CNN which preserves the semantic structure of multi-label images. The CNN model with listwise ranking supervision is used to jointly learn feature representations and mappings from them to binary codes. The resulting optimization problem of nonsmooth and multivariate ranking measure is solved by using a ranking loss on a set of triplets as the surrogate loss, which makes stochastic gradient descent could be used to optimize model parameters effectively. Extensive experiments demonstrate that the proposed method outperforms other state-of-the-art hashing methods in terms of ranking quality.

\section*{Acknowledgments}

This work was supported by the National Basic Research Program of China (2012CB316300), National Natural Science Foundation of China (61175003, 61135002, 61420106015, U1435221), and CCF-Tencent Open Fund.

{\small
\bibliographystyle{ieee}
\bibliography{cvpr2015zf}
}

\end{document}